\documentclass[acmsmall, table, xcdraw]{acmart}

\usepackage{url}
\usepackage{blindtext, graphicx}
\usepackage{amsmath}
\usepackage{amsthm} 
\usepackage{amssymb}
\usepackage{listings}
\usepackage{algorithm}
\usepackage{csquotes}
\usepackage{enumitem}
\usepackage{algorithm, algpseudocode}
\usepackage{subfigure}
\usepackage{hyperref}
\usepackage{wrapfig}
\usepackage{verbatim}
\usepackage[disable]{todonotes}
\usepackage{balance}
\usepackage{pdflscape}
\usepackage{afterpage}
\usepackage{capt-of}
\usepackage{booktabs}

\usepackage{enumerate}

\newcommand{\rfp}[2][inline]{\todo[color=green!80, #1]{Ramon: {\small #2}}}
\newcommand{\frm}[2][inline]{\todo[color=red!80, #1]{Felipe: {\small #2}}}
\newcommand{\no}[2][inline]{\todo[color=blue!30, #1]{Nir: {\small #2}}}

\newtheorem{definition}{Definition}

\newtheorem{example}{Example}

\theoremstyle{plain}

\newcommand\pred[1]{\texttt{\fontsize{8.5}{8.5}\selectfont{#1}}}

\usepackage{amsmath}

\def\gets{:=}

\def\predset{\Psi}
\def\opset{\mathcal{O}}

\def\factset{\Sigma}
\def\actionset{\mathcal{A}}
\def\tuple #1{\langle #1\rangle}
\def\contactions{C}

\def\planningdomain{\Xi}
\def\plan{\pi}
\def\optimalplan{\plan^{*}}
\def\optimalplans{\Pi^{*}}

\DeclareMathOperator*{\argmax}{arg\,max}

\def\state{s}

\usepackage{listings}
\def\BibTeX{{\rm B\kern-.05em{\sc i\kern-.025em b}\kern-.08emT\kern-.1667em\lower.7ex\hbox{E}\kern-.125emX}}
    
\begin{document}

\title{Using Sub-Optimal Plan Detection to Identify Commitment Abandonment in Discrete Environments}

\author{Ramon Fraga Pereira}
\email{ramon.pereira@edu.pucrs.br}
\affiliation{
  \institution{School of Technology, PUCRS}
  \city{Porto Alegre}
  \country{Brazil}
}

\author{Nir Oren}
\email{n.oren@abdn.ac.uk}
\affiliation{
  \institution{Department of Computing Science, University of Aberdeen}
  \city{Aberdeen}
  \country{Scotland}
}

\author{Felipe Meneguzzi}
\email{felipe.meneguzzi@pucrs.br}
\affiliation{
  \institution{School of Technology, PUCRS}
  \city{Porto Alegre}
  \country{Brazil}
}

\renewcommand{\shortauthors}{Pereira, Oren, and Meneguzzi}

\begin{abstract}

Assessing whether an agent has abandoned a goal or is actively pursuing it is important when multiple agents are trying to achieve joint goals, or when agents commit to achieving goals for each other. 
Making such a determination for a single goal by observing only plan traces is not trivial as agents often deviate from optimal plans for various reasons, including the pursuit of multiple goals or the inability to act optimally.
In this article, we develop an approach based on domain independent heuristics from automated planning, landmarks, and fact partitions to identify sub-optimal action steps --- with respect to a plan --- within a fully observable plan execution trace. 
Such capability is very important in domains where multiple agents cooperate and delegate tasks among themselves, e.g. through \emph{social commitments}, and need to ensure that a delegating agent can infer whether or not another agent is actually progressing towards a delegated task. 
We demonstrate how a creditor can use our technique to determine --- by observing a trace --- whether a debtor is honouring a commitment. 
We empirically show, for a number of representative domains, that our approach infers sub-optimal action steps with very high accuracy and detects commitment abandonment in nearly all cases.\footnote{Preliminary versions of parts of this article appeared as a 2 page extended-abstract \cite{Pereira:2017:DCA:3091282.3091404}, and a workshop paper \cite{PereiraOrenMeneguzzi_PAIR2017}. This article expands the problem formulation and formalisation; the descriptions and discussion of the heuristics and their implications for the technique; working examples and explanations; and experimentation.}

\end{abstract}


\maketitle

\section{Introduction}
\label{section:Introduction}

Autonomous agents generate and execute plans in pursuit of goals. Rationality would require such agents to execute plans which are --- in some sense --- \textit{optimal}. However, an agent may execute additional actions that are not part of an optimal plan due to factors including indecision, an imperfect planning mechanism, interleaving concurrent plans for multiple goals, and, in the most extreme case, goal or plan abandonment. 
sub-optimal execution of a plan or abandonment of a goal is not a major problem for an individual agent acting on its own, however, when agents work together and delegate goal achievement to each other, they need to be able to monitor and detect when an agent committed to acting on its behalf fails to comply with such commitment. 
Thus, determining whether observed actions are sub-optimal is often important, especially when goal delegation has taken place; where one agent is obliged or committed to achieve a goal; or where agents are coordinating plan execution in pursuit of a joint goal. 
In all these cases, determining that an agent is acting sub-optimally allows other agents to re-plan, notify or warn the agent, apply sanctions, or otherwise mitigate against the effects of the failure to achieve a certain state-of-affairs in a timely fashion.

Consider, for example, a situation where agents (such as trucks and airplanes) work together to deliver items to various destinations. 
By making commitments to each other, they are able to come up with a joint plan to perform item delivery. 
Clearly, all agents are interested in monitoring each other for deviations, enabling them to replan or impose sanctions if one of their partners begins behaving in an unexpected manner. 
Identifying such deviations involves monitoring how the partners are executing their part of the plan, and from this, determining if they are still committed to achieving it. 
While one could determine whether it is no longer possible for a partner to achieve the joint plan (e.g., if a truck can no longer deliver an item on time), it is useful to detect deviations earlier (e.g., detecting if the truck is moving away rather than towards its destination).

In this work, we address this problem of monitoring plan execution by detecting which steps in a plan are sub-optimal, that is, not contributing towards the agent's goal. 
By using a threshold on the number of sub-optimal actions, we can decide whether an agent has abandoned a monitored goal.
The use of the threshold value gives some flexibility to the observed agent to execute/perform actions that are part of other plans (e.g., consider that the agent has other goals to achieve).
Our contribution is twofold: first, we develop efficient techniques to compute whether a plan is sub-optimal and which actions in this plan are sub-optimal; second we leverage this technique to identify whether an agent is individually committed to achieve a particular goal, allowing us to identify whether this agent will honour a social commitment~\cite{AAAI-Commit-08}. 

The techniques we develop exploit domain-independent heuristics~\cite{AutomatedPlanning_Book2011}, planning landmarks~\cite{Hoffmann2004_OrderedLandmarks}, and fact partitions~\cite{PattisonGoalRecognition_2010} to monitor plan optimality and goal achievability (Section~\ref{section:PlanOptimalityMonitoring}). 
We assume that during plan execution, all actions performed by an agent are visible, and that a monitored goal and a domain theory (in a planning language) are available. We then evaluate the optimality of plan steps (i.e., actions) in two ways.
First, we estimate a distance (using any domain-independent heuristic) to the monitored goal at each step, analysing possible deviations. 
Second, we evaluate how each observation contributes toward a goal by analysing how they diminish their estimated distance to a sequence of states that must be achieved for the goal to eventually be achieved, also referred to as \emph{landmarks} \cite{Hoffmann2004_OrderedLandmarks}. 
With this information we can infer which observed actions are (probably) not part of an optimal plan. 
Our optimality monitoring approach can be contrasted with previous work on detecting whether a plan being executed aims to a monitored goal~\cite{FritzICAPS2007}. The work of Fritz and McIlraith~\cite{FritzICAPS2007} relies on a complex logical formalism and focused on extraneous events rather than directly on an agent's behaviour. We formalise the problem of \textit{commitment abandonment} detecting and the relation of an individual commitment to a plan in 
Section~\ref{section:CommitmentAbandonment}, using our plan optimality monitoring approach to detect whether an agent has abandoned a social commitment. This allows the creditor (observer) to ascertain at runtime whether, and when, the debtor fails to honour the commitment at the agreed upon quality. 

Experiments over several planning domains (Section~\ref{section:ExperimentsEvaluation}) show that our approaches yield high accuracy at low computational cost to detect sub-optimal actions (i.e., which actions do not contribute to achieve a monitored goal), and can, in nearly all evaluated cases, detect whether a debtor agent has abandoned a commitment. 

\section{Background}
\label{section:Background}

In this section, we review essential background on automated planning terminology, domain-independent heuristics, landmarks, and fact partitioning. 

\subsection{Planning}
\label{subsection:Planning}

Planning is the problem of finding a sequence of actions (i.e., a plan) that achieves a particular goal from an initial state. 
We adopt the terminology of Ghallab~et al.~\cite{AutomatedPlanning_Book2011} to represent planning domains and instances (also called planning  problems) in Definitions~\ref{def:State}--\ref{def:Plan}. We begin by considering \emph{states}, built up of predicates, which describe the environment at a moment in time.

\begin{definition} [\textbf{Predicates and State}]\label{def:State}
A predicate is denoted by an n-ary predicate symbol $p$ applied to a sequence of zero or more terms ($\tau_1$, $\tau_2$, ..., $\tau_n$) -- terms are either constants or variables.
We denote the set of all possible predicates as $\predset$.
We denote $\factset$ as the set of facts, which comprise all grounded predicates in both their positive or negated forms, as well as constants for truth ($\top$) and falsehood ($\bot$).
A state is a finite set of positive grounded predicates, representing logical values that are true in the state.
\end{definition}

Planning domains describe the environment's properties and dynamics through operators, which use a first-order representation to define schemata for state-modification actions according to Definitions~\ref{def:Operator}--\ref{def:Plan}.

\begin{definition} [\textbf{Operators and Actions}]\label{def:Operator}
An operator $a$ is a triple $\langle$\textit{name}($a$), \textit{pre}($a$), \textit{eff}($a$)$\rangle$ where \textit{name}($a$) is the description or \emph{signature} of $a$; \textit{pre}($a$) are its preconditions --- the set of predicates that must exist in the current state for $a$ to be executed; \textit{eff}($a$) represents the effects of $a$ which modify the current state. Effects are split into \textit{eff}($a$)$^+$ (i.e., an add-list of positive predicates) and \textit{eff}($a$)$^-$ (i.e., a delete-list of negated predicates).
An action is then a grounded operator instantiated over its free variables. The set of all operators is denoted by $\opset$, and the set of all possible actions is $\actionset$.
\end{definition}


We say an action $a$ is applicable to a state $S$ if and only if $S \models \mathit{pre}(a)$. This action generates a new state $S' := (S \cup \mathit{eff}(a)^{+})/\mathit{eff}(a)^{-}$. The application of an applicable action is captured through a function $\gamma(S,A)$ as follows.

$$\gamma(S,A) = \begin{cases}
	(S \cup \mathit{eff}(a)^{+})/\mathit{eff}(a)^{-} & \text{if}~S \models \mathit{pre}(a)\\
    \bot & \text{otherwise}
\end{cases}$$

\begin{definition}[\textbf{Planning Domain}]\label{def:PlanningDomain}
A planning domain definition $\Xi$ is represented by a pair $\langle \Sigma, \mathcal{A} \rangle$,  and consists of a finite set of grounded facts $\Sigma$ (e.g., environment properties) and a finite set of grounded actions $\mathcal{A}$.
\end{definition}

A \textit{planning instance} comprises both a \textit{planning domain} and the elements of a planning problem, describing a finite set of \textit{objects} of the environment, the \textit{initial state}, and the \textit{goal state} which an agent wishes to achieve.

\begin{definition} [\textbf{Planning Instance}]\label{def:PlanningInstance}
A planning instance is a triple $\Pi=\langle \Xi, \mathcal{I}, G\rangle$, where $\Xi =  \langle \Sigma, \mathcal{A}\rangle$ is a planning domain definition; $\mathcal{I} \subseteq \Sigma$ is the initial state specification, defined by specifying the value for all facts in the initial state; and $G \subseteq \Sigma$ is the goal state specification, which represents a desired subset of facts to be achieved.
\end{definition}

\begin{definition} [\textbf{Plan}]\label{def:Plan}
Let $\Pi=\langle \langle \Sigma, \mathcal{A}\rangle, \mathcal{I}, G\rangle$ be a planning instance. 
A plan $\pi$ for $\Pi$ is a sequence of applicable actions $[a_1, a_2, ..., a_n]$ (where $a_i \in \mathcal{A}$) that modifies the initial state $\mathcal{I}$ into one in which the goal state $G$ holds by the successive (ordered) execution of actions in a plan $\pi$, such that the preconditions of actions $[a_1, a_2, ..., a_n]$ are satisfied throughout the execution of the plan $\pi$, i.e., $\gamma(\gamma(\gamma(...\gamma(\mathcal{I},a_1)...),a_{n_1}), a_n) \models G$.
\end{definition}



Planners often exploit heuristics which estimate the cost to achieve a specific goal from some state~\cite{AutomatedPlanning_Book2011}. 
In this work, and as done by many other classical planners, we consider that all actions have equal cost, making the cost of a plan equal to its length.
When a heuristic never overestimates the cost to achieve a goal, it is called \emph{admissible} and guarantees optimal plans when used with  certain planning algorithms. 
In this work, we use both admissible and inadmissible domain-independent heuristics for estimating the distance to a monitored goal.


\subsection{Landmarks}
\label{subsection:Landmarks}

Planning landmarks are necessary properties (or actions) that must be true (or executed) at some point in every valid plan (c.f. Definition~\ref{def:Plan}) to achieve a particular goal from an initial state. 
Hoffman~et al.~\cite{Hoffmann2004_OrderedLandmarks} define landmarks as follows.

\begin{definition}[\textbf{Fact Landmarks}]
Given a planning instance $\Pi=\langle \Xi, \mathcal{I}, G\rangle$, a formula $L$ is a landmark in $\Pi$ iff $L$ is true at some point along all valid plans that achieve a goal state $G$ from an initial state $\mathcal{I}$. 
\end{definition}

\begin{figure}[t!]
  \centering
  \includegraphics[width=0.7\linewidth]{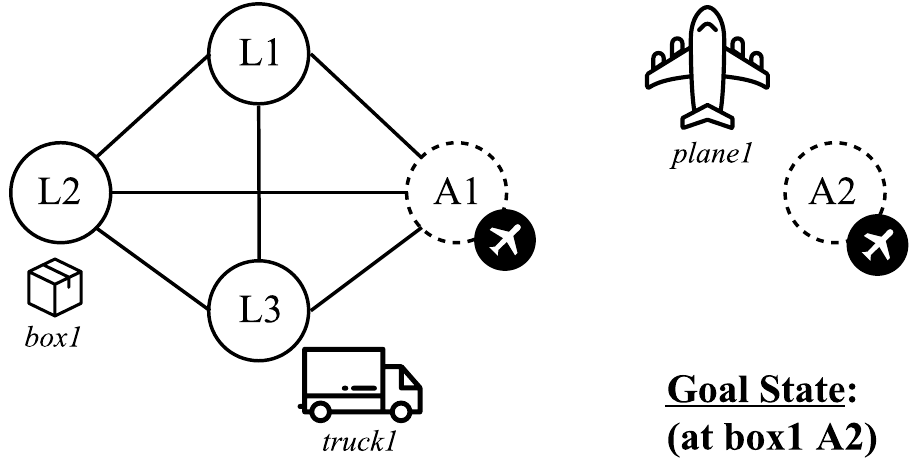}
  \caption{\textsc{Logistics} problem example.}
  \label{fig:logisticsExample}
\end{figure}

\begin{figure}[ht]
  \centering
  \includegraphics[width=0.7\linewidth]{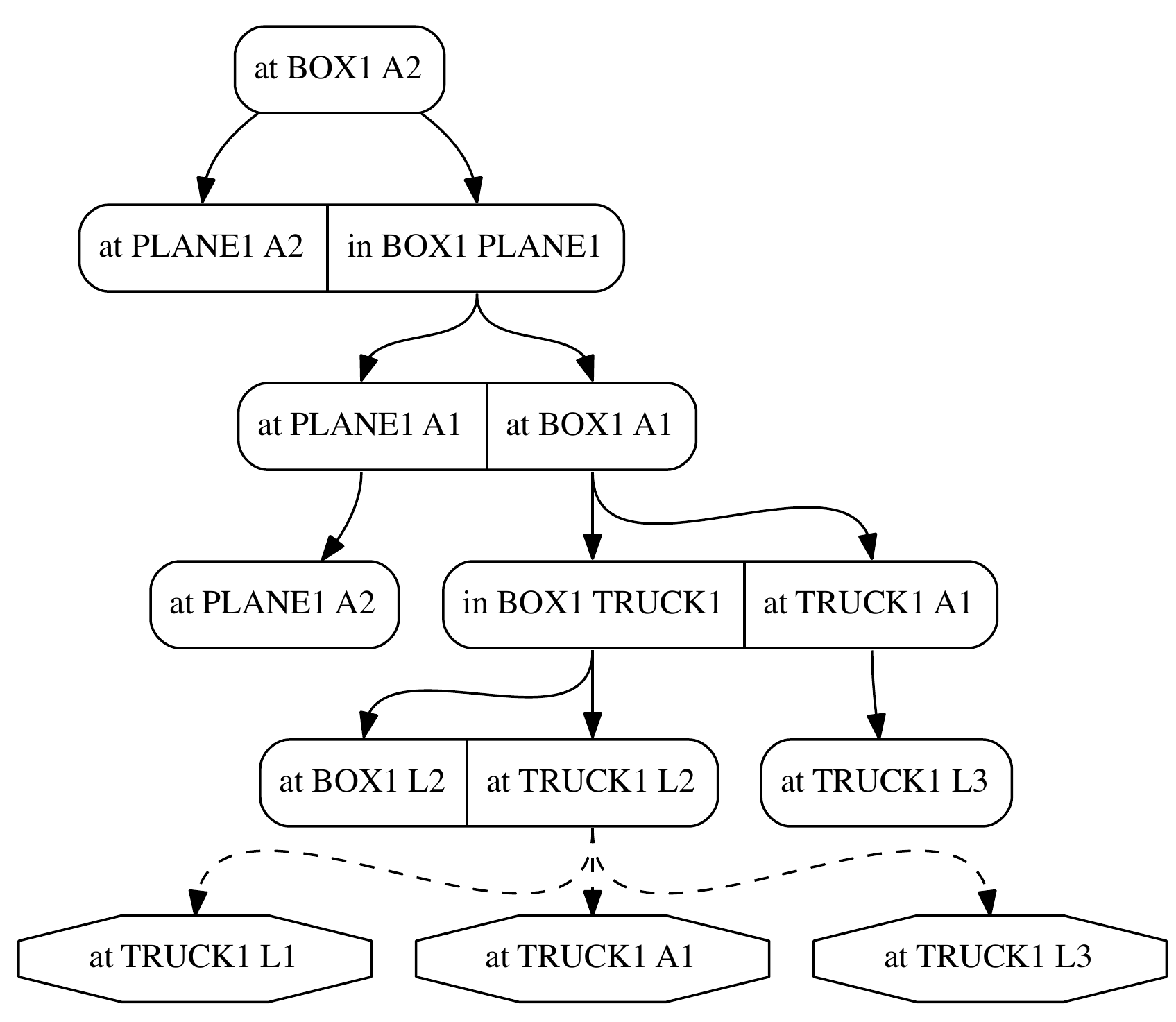}
  \caption{Ordered fact landmarks extracted from the \textsc{Logistics}  example from Figure~\ref{fig:logisticsExample}. Fact landmarks that must be true together are represented by connected boxes and represent conjunctive landmarks. Disjunctive landmarks are represented by octagonal boxes connected by dashed lines.}
  \label{fig:LogisticsLandmarks}
\end{figure}

Hoffmann~et al.~\cite{Hoffmann2004_OrderedLandmarks} describe both conjunctive and disjunctive landmarks. 
A conjunctive landmark is a set of facts that must be true together at some state in every valid plan to achieve a goal. 
A disjunctive landmark is a set of facts in which one of facts must be true at some state in every valid plan to achieve a goal. 
Landmarks are often partially ordered by their pre-requisite dependencies. 
The process of landmark extraction both identifies conjunctive and disjunctive landmarks, and determines the temporal ordering between them (i.e., identifies which landmark occurs before which). As an example of landmarks and their orderings, consider an instance of the \textsc{Logistics}\footnote{\textsc{Logistics} is a domain (adapted from~\cite{AIPS_98}) that consists of airplanes and trucks transporting packages between locations (e.g., airports and cities).} planning problem shown in Figure~\ref{fig:logisticsExample}. 
This example shows two cities: the city on the left contains locations L1 through L3 and an airport (A1), and the city on the right contains another airport (A2). 
The goal within this example is to transport an item (box1) from location L2 to location A2. Listing~\ref{lst:LogisticsFactLandmarks} shows the resulting fact landmarks, while Figure~\ref{fig:LogisticsLandmarks} shows their ordering (edges show that a source formula must hold after its target). Note that a goal is considered to be a conjunctive landmark. From Figure \ref{fig:LogisticsLandmarks}, we see that the second landmark (stating that  any valid plan must have box1 within plane1, and that plane1 must be at airport A2) must occur before the goal is achieved, and that before the box is within the plane, plane1 must be at A1, and so must box1.

\begin{lstlisting}[float=!tb,caption={Fact landmarks (conjunctive and disjunctive) extracted from the \textsc{Logistics} example.},label={lst:LogisticsFactLandmarks},basicstyle=\ttfamily\footnotesize]
Fact Landmarks:
(and (at BOX1 A2))
(and (at PLANE1 A2) (in BOX1 PLANE1))
(and (at PLANE1 A1) (at BOX1 A1))
(and (at PLANE1 A2))
(and (at TRUCK1 L3))(and (at TRUCK1 L3))
(and (in BOX1 TRUCK1) (at TRUCK1 A1))
(and (at BOX1 L2) (at TRUCK1 L2))
(or (at TRUCK1 L1) (at TRUCK1 A1) (at TRUCK1 L3))
\end{lstlisting}

Whereas in automated planning the concept of landmarks is used to build heuristics~\cite{LandmarksRichter_2008,KeyderRH_ECAI10_Landmarks,DOMSHLAK_AIJ2012_Landmarks} and as a fundamental part of planning algorithms~\cite{RichterLPG_2010,HelmertFastDownward_2011}, in this work, we propose to use landmarks to monitor an agent's plan execution and detect sub-optimal actions in a plan. 
Intuitively, we use landmarks as waypoints (or stepping stones) to monitor what states (or actions) an agent cannot avoid while seeking to achieve its goal.

In the planning literature there are several algorithms to extract landmarks and their orderings~\cite{ICAPS03_DC_ZhGi,LandmarksRichter_2008,KeyderRH_ECAI10_Landmarks}, and in this work, for extracting landmarks from planning instances we use the algorithm of Hoffmann~et al.~\cite{Hoffmann2004_OrderedLandmarks}.  
We note that many landmark extraction techniques, including that of Hoffmann~et al.~\cite{Hoffmann2004_OrderedLandmarks}, have the potential to infer incorrect landmark orderings, which can lead to problems if the optimality monitoring problem relies on the ordering information to make inferences. 
Nevertheless, our empirical evaluation shows that landmark orderings do not affect detection performance in our experimental dataset, and we discuss landmark orderings later in the article.


\subsection{Fact Partitioning}
\label{sec:fact-partitioning}


To perform goal recognition, Pattison and Long~\cite{PattisonGoalRecognition_2010} classify facts into mutually exclusive partitions so as to infer whether certain observations are likely to be goals.
Their classification relies on the fact that --- in some planning domains --- predicates may provide additional information that can be extracted by analysing preconditions and effects in operator definitions. 
Given a set of candidate goals, we use this classification to infer whether certain observations are consistent with a particular goal. If an inconsistency is detected, we can eliminate the candidate goal. Pattison and Long's classification can be formalised as follows.


\begin{definition} [\textbf{Strictly Activating}] \label{def:strictlyActivating}
A fact $f$ is strictly activating if $f \in \mathcal{I}$ and $\forall a \in \mathcal{A}$, $f \notin \textit{eff}(a)^+ \cup \textit{eff}(a)^-$. Furthermore, $\exists a \in \mathcal{A}$, such that $f \in$ \textit{pre}($a$).
\end{definition}

\begin{definition} [\textbf{Unstable Activating}] \label{def:unstableActivating}
A fact $f$ is unstable activating if $f \in \mathcal{I}$ and $\forall a \in \mathcal{A}$, $f \notin \textit{eff}(a)^+$ and $\exists a,b \in \mathcal{A}, f \in \textit{pre}(a)$ and $f \in \textit{eff}(b)^-$.
\end{definition}

\begin{definition} [\textbf{Strictly Terminal}] \label{def:strictlyTerminal}
A fact $f$ is strictly terminal if $\exists a \in \mathcal{A}$, such that $f \in \textit{eff}(a)^+$ and $\forall a \in \mathcal{A}$, $f \notin \textit{pre}(a)$ and $f \notin \textit{eff}(a)^-$.
\end{definition}

A \textit{Strictly Activating} fact (Definition~\ref{def:strictlyActivating}) appears as a precondition, and does not appear as an add or delete effect in an operator definition. 
Unless defined in the initial state, such a fact can never be added or deleted by an operator.
An \textit{Unstable Activating} fact (Definition~\ref{def:unstableActivating}) appears as both a precondition and a delete effect in two operator definitions, so once deleted, this fact cannot be re-achieved. 
The deletion of an unstable activating fact may prevent a plan execution from achieving a goal.
A \textit{Strictly Terminal} fact (Definition~\ref{def:strictlyTerminal}) does not appear as a precondition of any operator definition, and once added, cannot be deleted. 
For some planning domains, this kind of fact is most likely to be in the set of goal facts, because once added in the current state, it cannot be deleted, and remains true until the final state.

The algorithm we describe in the next section utilises fact partitioning to improve its performance by determining whether a goal is or is not achievable. We note that detecting fact partitions depends on the planning domain definition, and more specifically, operator definition. For example, consider an \textit{Unstable Activating} fact. If an action deletes this fact from the current state, it cannot be re-achieved, and any goals which depend on this fact (i.e., for which it is a landmark) are unreachable. 
However, the presence or absence of fact partitions is highly domain dependent. For example, from the \textsc{Blocks-World}\footnote{\textsc{Blocks-World} is a classical planning domain where a set of stackable blocks must be re-assembled on a table~\cite[Chapter~2, Page~50]{AutomatedPlanning_Book2011} (also appeared in~\cite{AIPS_98}).} domain, it is not possible to extract any fact partitions, while the \textsc{Easy-IPC-Grid}\footnote{\textsc{Easy-IPC-Grid} is a domain that consists of an agent that moves in a grid using keys to open locked locations (adapted from~\cite{AIPS_98}).} domain contains \textit{Strictly Activating} and \textit{Unstable Activating} facts. Therefore, while our algorithm can exploit fact partitions, they are not required for the algorithm's operation.

\subsection{Commitments}

Commitments have been used in multi-agent systems to enable autonomous agents to communicate and coordinate successfully to achieve a particular goal \cite{Commitments_MeneguzziTS13,CommitmentsHTN_Telang2013,CommitmentsJaCaMo_BaldoniBCM15}. 
A commitment \texttt{C(DEBTOR, CREDITOR, antecedent, consequent)} formalises that the agent \texttt{DEBTOR} commits to  agent \texttt{CREDITOR} to bring about the \texttt{consequent} if the \texttt{antecedent} holds. 
Here, the antecedent and consequent conditions are conjunctions or disjunctions of events and possibly other commitments.  

In this work, we aim to monitor the \texttt{DEBTOR}'s behaviour (i.e., sequence of actions, a plan) to detect if this agent is individually committed to carrying out a plan to achieve the consequent for the \texttt{CREDITOR}. In  Section~\ref{subsection:DetectingCommitmentAbandonment}, we detail how we formalize commitments making an analogy to automated-planning, much like the work of Meneguzzi et al.~\cite{Commitments_MeneguzziTS13}, and how we detect commitment abandonment by combining the techniques developed in this article.

\section{Monitoring and Detecting Plan Optimality}
\label{section:PlanOptimalityMonitoring}

We now describe our plan optimality monitoring approach that uses landmarks, fact partitioning, and domain-independent heuristics. 
Intuitively, this approach aims to detect which actions in the execution of an agent plan do not contribute to the plan (\emph{sub-optimal actions}) for achieving the monitored goal. 
We begin by formalising the notion of plan optimality. 
Then, we describe a method that uses heuristics to estimate the distance to some monitored goal for every observed action in the plan execution, and infer whether there is any deviation in the observed plan to achieve the monitored goal. 
Following this, we develop a method that uses landmarks to anticipate what action the observed agent has to perform in the next observation to reduce the estimated distance to the next landmarks, and consequently to the monitored goal. Finally, we describe how plan optimality monitoring can be performed by bringing together these two previous methods. 

\subsection{Plan Optimality Monitoring Problem}
\label{subsection:OptimalityMonitoringProblem}

We define plan optimality monitoring as the process of monitoring the execution of a plan by an agent to solve a planning instance (Definition~\ref{def:PlanningInstance}) and detecting when the agent executes steps that deviate from any one of the optimal plans which exist for the planning instance~\cite{PereiraOrenMeneguzzi_PAIR2017}.  
Formally, we want to detect when the observed agent fails to execute one of the optimal plans of Definition~\ref{def:OptimalPlan}, and instead executes any of the (possibly infinite) number of valid sub-optimal plans.

\begin{definition}[\textbf{Optimal Plan}]\label{def:OptimalPlan}
Let $\pi = [a_1,...a_n]$ be a plan with length $|\pi|=n$ for a domain $\Pi$, we say $\pi$ is optimal, also written as $\pi^{*}$ if there exists no other plan $\pi^{<}$ such that $|\pi^{<}| < |\pi^{*}|$.
\end{definition}


When an agent executes a plan in an environment it is called \emph{plan execution}, formally defined in Definition~\ref{def:PlanExecution}, and this work, such an execution generates an \emph{observation sequence}, formalised in Definition~\ref{def:Observation}. 
In fully observable environments, there is a one-to-one correspondence between the actions in a plan and observations. 

\begin{definition} [\textbf{Plan Execution}]\label{def:PlanExecution}
A plan execution $\pi_{E}$ is the execution of a sequence of applicable actions (i.e., a plan $\pi = [a_1,...a_n]$) from an initial state $\mathcal{I}$ to a particular state. 
A plan execution $\pi_{E}$ can be either optimal or sub-optimal depending on the plan.
\end{definition}

\begin{definition} [\textbf{Observation Sequence}]\label{def:Observation}
Let $O$ be a sequence $[o_1, o_2, ..., o_n]$ of observations of a plan's execution with each observation $o_i \in O$ such that $o_{i}=name(a)$ for some $a \in \mathcal{A}$, i.e. the name of some instantiated action in the set of actions in a domain definition $\Xi$.
\end{definition}

Intuitively, given a planning instance, we want to detect exactly which actions and their sequence in the plan execution do not contribute towards the monitored goal in a planning instance. 
We formally define the task of plan optimality monitoring in Definition~\ref{def:OptimalityMonitoring} and note that, for this task, we consider that we always have full observability, so we observe all actions during a plan execution.

\begin{definition} [\textbf{Plan Optimality Monitoring Problem}]\label{def:OptimalityMonitoring}
A plan optimality monitoring problem is a tuple $T_{\pi^{*}} = \langle \Pi, O \rangle$, where (i) $\Pi = \langle \planningdomain, \mathcal{I}, G \rangle$ is a planning instance with domain definition $\planningdomain = \langle \Sigma, \mathcal{A} \rangle$, an initial state $\mathcal{I}$, and goal $G$; 
and (ii)  $O = \langle o_1, o_2, ..., o_n \rangle$ is an observation sequence of the plan execution.
\end{definition}




In solving the plan optimality monitoring problem, we seek those actions in the observation sequence that do not contribute to the achievement of the monitored goal $G$ from the initial state $\mathcal{I}$. 
To do so, we formalise \emph{contributing actions} in Definition~\ref{def:contributingAction}. 
Informally, a \emph{contributing action} is identified by selecting, from the set of optimal plans $\optimalplans$, that which maximally matches the sequence of observations, and choosing actions from this plan that lead to the goal state. 
Thus, the \emph{non-contributing actions} are those that are found in the observations but which diverge from the optimal plan.  
More formally, the sub-optimal actions in $O$ are the actions in $O - \optimalplan$. 

\begin{definition} [\textbf{Contributing Action}]\label{def:contributingAction}
Let $T_{\pi^{*}} = \langle \tuple{\planningdomain, \mathcal{I}, G}, O \rangle$ be a plan optimality monitoring problem with an associated set of optimal plans $\optimalplans$. 
We define the \emph{sequence} of contributing actions $\contactions$ with regards to a plan $\plan$ as follows. 

$$\contactions(\tuple{\tuple{\planningdomain, \mathcal{I}, G}, O},\plan) = 
\begin{cases} 
	\tuple{ o_{1} } + \contactions(\tuple{\tuple{\planningdomain, \gamma(\mathcal{I},o_1), G}, \tuple{o_2,..., o_n}}) & \text{if~} o_{1} \in \plan \\
	\contactions(\tuple{\tuple{\planningdomain, \gamma(\mathcal{I},o_1), G}, \tuple{o_2,..., o_n}}) & \text{if~} o_{1} \not\in \plan \\
	\tuple{~} & otherwise \\
\end{cases}$$

Here, $+$ denotes the concatenation of two sequences. 
We say $O$ is consistent with an optimal plan $\optimalplan \in \optimalplans$ if and only if $\optimalplan = \argmax_{\plan \in \optimalplans}|C(\langle \tuple{\planningdomain, \mathcal{I}, G}, O \rangle, \plan)|$ and note that there may be multiple such plans for a given domain. 
\end{definition}

\no{Another experiment - at how early in the observation sequence can we detect plan deviation? This can be thought of as online rather than offline plan optimality monitoring.}
\rfp{We can do it in a online fashion, it is quite easy, we just have to do it incrementally as we see an observation, and return if the observed action is actually an sub-optimal step or not.}
\rfp{I think that we must emphasise this: our approaches can perform both in offline and online fashion.}

Note that the definition of contributing action does allow us to solve this problem in an online fashion, that is, to evaluate observations as the observer perceives them and computing whether these actions are optimal or not. 
\frm{Discuss the possible wobbles on what is optimal when done online.}
\frm{$o_1$ needs to be the next action in any of the plans in $\pi^*$}

\begin{example}\label{exemp:PlanOptimalityMonitoring}
Consider the \textsc{Logistics} example in {\normalfont Figure~\ref{fig:logisticsExample}}, which shows a planning problem with two cities: the city on the left that contains four locations: L1 through L3; and the city on the right that contains only the location A2. 
Locations A1 and A2 are airports. 
The goal of this example is to transport the box located at location L2 to location A2. 
From this planning problem, {\normalfont Table~\ref{tab:PlanAbandomentLogisticsExample}} shows the execution of two possible plans: an optimal and a sub-optimal. 
In the sub-optimal plan execution, grey actions are those that do not contribute to achieve the goal, i.e., sub-optimal actions.
Light-grey actions are those that must be taken to undo the non-contributing actions and thus ultimately achieve the goal.
\end{example}

\begin{table*}[!ht]
\footnotesize
\centering
\begin{tabular}{lllll}
\multicolumn{1}{c}{\bf Optimal Plan}         &  & \multicolumn{1}{c}{\bf Sub-Optimal Plan}                                       &  &  \\
                                         &  &                                                                            &  &  \\
0 - \texttt{(drive TRUCK1 L3 L2 CITY1)}              &  & 0 - \texttt{(drive TRUCK1 L3 L2 CITY1)}                                                &  &  \\
1 - \texttt{(loadTruck BOX1 TRUCK1 L2)}              &  & 1 - \texttt{(loadTruck BOX1 TRUCK1 L2)}                                                &  &  \\
2 - \texttt{(drive TRUCK1 L2 A1 CITY1)}      &  & \cellcolor[HTML]{9B9B9B}2 - \texttt{(unloadTruck BOX1 TRUCK1 L2)}                      &  &  \\
3 - \texttt{(unloadTruck BOX1 TRUCK1 A1)}    &  & \cellcolor[HTML]{9B9B9B}3 - \texttt{(drive TRUCK1 L2 L1 CITY1)}                        &  &  \\
4 - \texttt{(fly PLANE1 A2 A1)}      &  & \cellcolor[HTML]{C0C0C0}{\color[HTML]{333333} 4 - \texttt{(drive TRUCK1 L1 L2 CITY1)}} &  &  \\
5 - \texttt{(loadAirPlane BOX1 PLANE1 A1)}   &  & \cellcolor[HTML]{C0C0C0}{\color[HTML]{333333} 5 - \texttt{(loadTruck BOX1 TRUCK1 L2)}} &  &  \\
6 - \texttt{(fly PLANE1 A1 A2)}      &  & 6 - \texttt{(drive TRUCK1 L2 A1 CITY1)}                                        &  &  \\
7 - \texttt{(unloadAirplane BOX1 PLANE1 A2)} &  & 7 - \texttt{(unloadTruck BOX1 TRUCK1 A1)}                                      &  &  \\
                                         &  & 8 - \texttt{(fly PLANE1 A2 A1)}                                        &  &  \\
                                         &  & 9 - \texttt{(loadAirPlane BOX1 PLANE1 A1)}                                     &  &  \\
                                         &  & 10 - \texttt{(fly PLANE A1 A2)}                                       &  &  \\
                                         &  & 11 - \texttt{(unloadAirplane BOX1 PLANE1 A2)}                                  &  &
\end{tabular}
\caption{Plan Optimality Monitoring example.}
\label{tab:PlanAbandomentLogisticsExample}
\end{table*}

\subsection{Analysing Plan Execution Deviation}
\label{subsection:AnalysingPlanDeviation}

We now develop a method that analyses a plan execution to identify plan deviation for achieving a goal state from an initial state.
To analyse possible plan execution deviation in an observation sequence (Definition~\ref{def:Observation}), we compute the estimated distance to the monitored goal for every state resulting from the execution of an observed action. 
Note that we assume full plan observability in the sense that no actions are missing from the observations from the initial state up to a time point. 
For example, if a optimal plan to goal $G$ has $10$ action steps, and we observe just $4$ actions in the observation sequence, then $O_{G}$ = $[o_1, o_2, o_3, o_4]$ corresponds exactly to the four first actions in the plan. 

\begin{figure}[ht]
\begin{center}
    \includegraphics[width=0.65\linewidth]{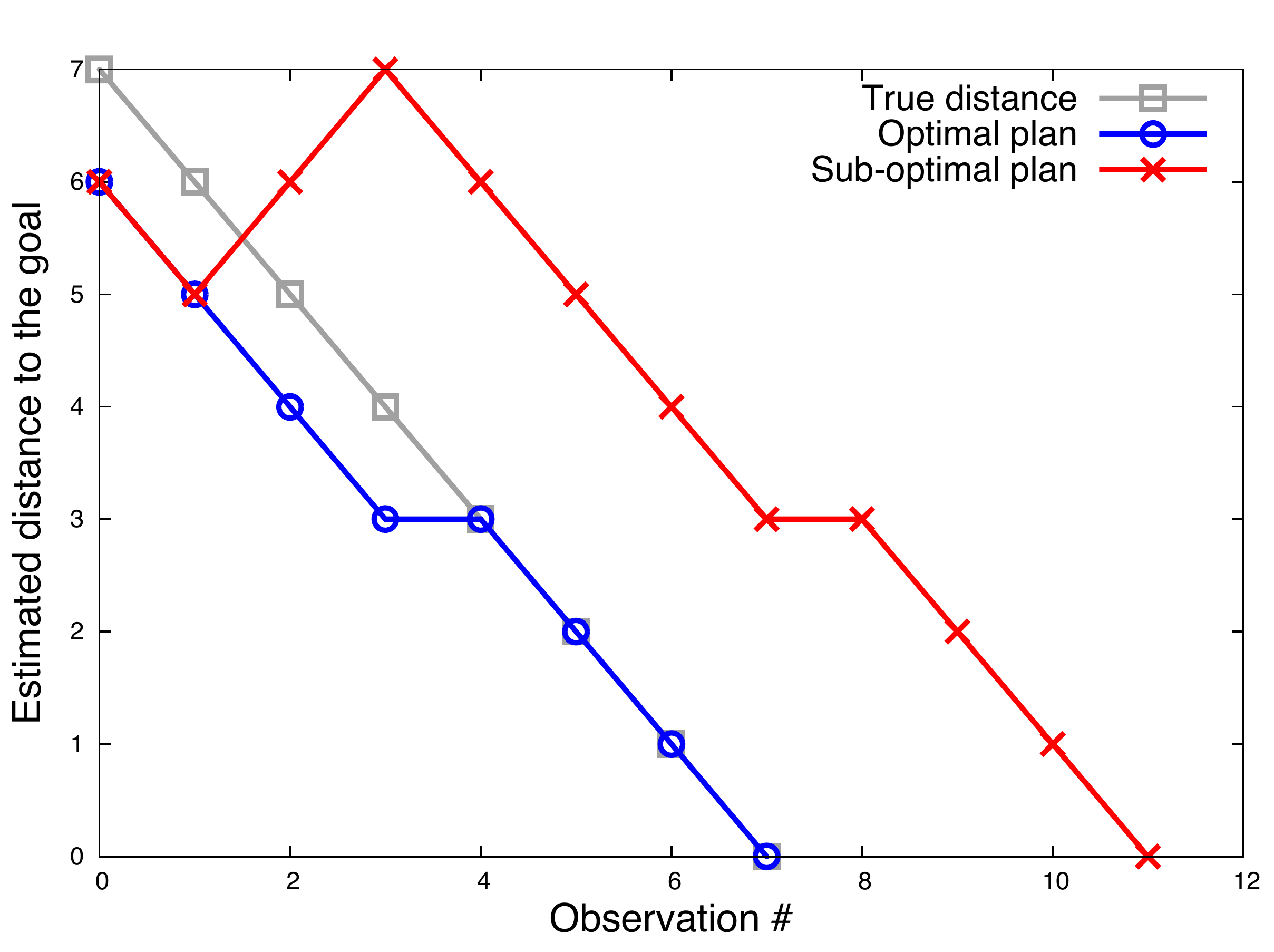}
    \caption{Plan execution deviation example using the \textsc{Fast-Forward} heuristic.}
    \label{fig:PlanExecutionHeuristics}
\end{center}
\end{figure}

Given a state $s$, a heuristic $h$ returns an estimated distance $h(s)$ to the goal state~\cite{BonetGeffner_PlanningAsHeuristic_2001}. 
For example, consider the states $s_{i}$ and $s_{i-1}$ and the observed action $o_i$ at the step $i$. 
If the observed action $o_i$ at step $i$ transitions the system to state $s_i$, we consider a deviation from a plan to occur if $h(s_{i-1}) < h(s_i)$. 
Such deviations can arise for a variety of reasons including concurrent or interleaved plan execution by the agent (e.g., in an attempt to achieve multiple goals simultaneously); non-optimal plan selection (e.g., due to bounded rationality); and incorrect estimates by the heuristic.  
The up-tick shown in Figure~\ref{fig:PlanExecutionHeuristics} illustrates a deviation detected using the \textsc{Fast-Forward} heuristic~\cite{FFHoffmann_2001} for two different plan executions. 
These two plan executions (an optimal plan -- points denoted as circles, and a sub-optimal plan -- points denoted as crosses) are plans that achieve the goal state from the initial state in Figure~\ref{fig:logisticsExample}. 
During the execution of the sub-optimal plan, deviations occur for actions leading at the observation time 2 and 3. 
By analysing this plan deviation, we conclude that these actions do not contribute to achieve the goal because they increase the estimated distance to the goal state. 
However, given the potential large deviations from sub-optimal plans combined with the varying ways in which heuristics can inaccurately measure the distance towards a goal, we cannot rely exclusively on deviations from the heuristic to detect sub-optimal actions. 


Thus, since heuristics may be inaccurate, we use landmarks to build a further condition of sub-optimality, predicting actions that achieve next landmarks, and consequently the monitored goal state.

%
%

\subsection{Predicting Upcoming Actions via Landmarks}
\label{subsection:PredictingUpcomingActions}

Ordered landmarks~\cite{Hoffmann2004_OrderedLandmarks} effectively provide way-points towards a monitored goal from an initial state, and based on these way-points we can infer what cannot be avoided on the way to achieving such monitored goal. Note that the initial and goal state are themselves landmarks, as all plans begin and terminate in these states.
Since all plans should pass through a landmark, we can exploit their presence to predict what actions might be executed next, either to reach the ordered landmarks, or to progress towards a monitored goal. 
We use such predictions to check the set of observed actions of a plan execution in order to determine which actions do not contribute to achieve the monitored goal~\cite{PereiraOrenMeneguzzi_PAIR2017}. The use of lookahead actions is quite common in AI Planning and search~\cite{Vidal_ICAPS_2004,Lipovetzky_PROBES_ICAPS11}, however, in this article, we use such look ahead (or predictions) to verify if the plan execution is progressing towards the monitored goal without deviating.
We formalise this in Algorithm~\ref{alg:PredictiNextActions}.

To predict which actions could reasonably be executed in the next observation and to minimise the accumulated discrepancies due to the imprecise nature of the heuristics, our algorithm identifies the closest landmarks by estimating the distance to the landmarks from the current state. 
Our approach uses an admissible domain-independent heuristic to estimate the distance to landmarks, namely the \textsc{Max-Heuristic} (or Max-Cost heuristic), which we denote as $h_{max}$. 
This heuristic, originally defined by \cite{BonetGeffner_PlanningAsHeuristic_2001} consists of computing the costs of achieving each individual literal $g_i$ in a goal $G$ as follows\footnote{Since we assume unit-cost actions, $cost(a)=1$}:
\begin{align*}
	h_{max}(s,G)    & = \max_{g_i \in G}h_{max}(s,g_i)\\
	h_{max}(s,g_i)  & = \begin{cases}
	0, & \text{if}~g_{i} \in s,\\
	\min\{h_{max}(s,a) | a \in \mathcal{A} \text{~and~} g_{i} \in \mathit{eff}(a)\}, & \text{otherwise;}
	\end{cases}	\\
	h_{max}(s,a) & = \mathit{cost}(a) + h_{max}(s,\mathit{pre}(a))
\end{align*}
%
We consider that the neighbouring fact landmarks are those that return estimated distance $h_{max}(s,l) = 0$ and $h_{max}(s,l) = 1$. We use these neighbouring landmarks to identify --- where possible --- a set of actions which should be executed next. The resulting algorithm (Algorithm \ref{alg:compute}) iterates over a set of ordered fact landmarks $\mathcal{L}$ (Line~\ref{alg:PredictiNextActions:for}), and, for each landmark $l$, the \textsc{Max-Heuristic} estimates the distance from the current state $s$ to $l$. 
If the estimated distance to landmark $l$ is $h_{max}(s,l)$ = 0 (Line~\ref{alg:PredictiNextActions:h0}), this means that landmark $l$ is in the current state, and the algorithm selects those actions that contain $l$ as a precondition, because these can be executed immediately (Line~\ref{alg:PredictiNextActions:now}). 
Otherwise, if the estimated distance to landmark $l$ is $h_{max}(s,l)$ = 1 (Line~\ref{alg:PredictiNextActions:h1}), this means that landmark $l$ can be reached by executing a single action, and the algorithm selects those actions that are applicable in the current state and contain $l$ as an effect (Line~\ref{alg:PredictiNextActions:next}). 
These actions are selected because they reduce the distance to the next landmark, and consequently to the monitored goal. 
Thus, we use the observed plan execution to estimate which actions do not contribute to achieve a goal. 
Example~\ref{exemp:predictingUpcomingActions} shows how Algorithm~\ref{alg:PredictiNextActions} predicts upcoming actions using landmarks.


\floatname{algorithm}{Algorithm}
\begin{algorithm}[ht]
    \caption{\label{alg:compute} Compute Upcoming Actions via Landmarks.}
	\begin{flushleft}
    \textbf{Parameters:} $\Xi = \langle \Sigma$, $\mathcal{A} \rangle$ \textit{planning domain}, $\state$ \textit{current state}, and $\mathcal{L}$ \textit{ordered fact landmarks}.
    \\\textbf{Output:} $\eta_{PActions}$ \textit{set of possible upcoming actions}.\label{alg:PredictiNextActions}
	\end{flushleft}
    \begin{algorithmic}[1]
        \Function{PredictUpcomingActions}{$\Xi$, $\state$, $\mathcal{L}$}
        		\State $\eta_{PActions} \gets \emptyset$ 
			\For{each fact landmark $l$ in $\mathcal{L}$} \label{alg:PredictiNextActions:for}
				\State $Al \gets \emptyset$
            		\If{$h_{max}(s,l) = 0$} \Comment{$h_{max}(s,l)$ \textit{estimates} $l$ \textit{from} $\state$.} \label{alg:PredictiNextActions:h0}
            			\State $Al \gets$ all $a$ in $\mathcal{A}$ s.t. $l \in \textit{pre}(a)$ \label{alg:PredictiNextActions:now}
				\ElsIf{$h_{max}(s,l) = 1$} \label{alg:PredictiNextActions:h1}
					\State $Al \gets$ all $a \in \mathcal{A}$ s.t $\textit{pre}(a) \in \state \land l \in \textit{eff}(a)^+$\label{alg:PredictiNextActions:next}
				\EndIf
				\State $\eta_{PActions}$ := $\eta_{PActions}$ $\cup$ $Al$
            \EndFor
            	\State \textbf{return} $\eta_{PActions}$
        \EndFunction
    \end{algorithmic}
\end{algorithm}

\begin{example}\label{exemp:predictingUpcomingActions}
Consider the \textsc{Logistics} problem in {\normalfont Figure~\ref{fig:logisticsExample}}. 
If the current state is the initial state, then the algorithm predicts upcoming actions that might be executed as the first observation in the plan execution. 
{\normalfont Table~\ref{tab:PredictorExample}} shows three columns; (1) the set of fact landmarks to achieve the goal from the initial state (Landmarks); (2) the estimated distance from the current state (i.e., initial state) to fact landmarks using $h_{max}(l)$; and (3) which applicable actions our algorithm predicts to be in the next observation (Upcoming Actions). 
From these landmarks, three of them have $h_{max}(s,l) = 0$, namely {\normalfont \pred{(at PLANE1 A1)}}, {\normalfont \pred{(at TRUCK1 L3)}}, and {\normalfont \pred{(at TRUCK1 L3)}}; and other three have $h_{max}(s,l) = 1$, namely {\normalfont \pred{(and (in BOX1 TRUCK1) (at TRUCK1 A1))}}, {\normalfont \pred{(at TRUCK1 L1)}}, and {\normalfont \pred{(at TRUCK1 A1)}}. 
Note that, there is no upcoming (predicted) actions for the fact landmarks for which the estimated distance is $h_{max}(s,l) = 1$, because there is no applicable action in the initial state to achieve these fact landmarks.
Thus, from the landmarks which have the estimated distance $h_{max}(s,l) = 0$, Algorithm~\ref{alg:PredictiNextActions} predicts two actions as the first expected observation: {\normalfont \pred{(fly PLANE1 A2 A1)}} or {\normalfont \pred{(drive TRUCK1 L3 L2 CITY1)}}.
These actions aim to reduce the distance to the next ordered landmarks, and consequently to the monitored goal.
\end{example}

\begin{table*}[h!]
\footnotesize
\centering
\begin{tabular}{lll}
\textbf{Landmarks} & $\mathit{h_{max}(s,l)}$ & \textbf{Predicted Actions} \\
\pred{(and (at BOX1 A2))}           		      &       7             &       -           \\
\pred{(and (at PLANE1 A2) (in BOX1 PLANE1))}      &		  6				&		-			\\
\pred{(and (at PLANE1 A1) (at BOX1 A2))}  &       5				&		-			\\
\pred{(and (at PLANE1 A2))}					  &		  0				&\pred{(fly PLANE1 A2 A1)} \\
\pred{(and (at TRUCK1 L3))}							  & 	  0				&\pred{(drive TRUCK1 L3 L2 CITY1)} \\
\pred{(and (in BOX1 TRUCK1) (at TRUCK AIRPORT-C))}	  & 	  3				& 		-			\\
\pred{(and (at BOX1 L2) (at TRUCK1 L2))}				  &		  1				&		-			\\
\pred{(or}											  & 					& 					\\
\pred{   (at TRUCK1 L1)}								  &		  1				&		-			\\
\pred{   (at TRUCK1 A1)}								  &		  1				&		-			\\
\pred{   (at TRUCK1 L3))}								  &		  0				&\pred{(drive TRUCK L3 L2 CITY1)}			\\

          &                    &                 
\end{tabular}
\caption{Predicted upcoming actions for the \textsc{Logistics} example in Figure~\ref{fig:logisticsExample}.}
\label{tab:PredictorExample}
\end{table*}

\subsection{Detecting Sub-Optimal Action Steps}
\label{subsection:Detectingsub-optimalSteps}

We now develop our approach to detect sub-optimal action steps~\cite{PereiraOrenMeneguzzi_PAIR2017}, bringing together the methods that were presented in Sections~\ref{subsection:AnalysingPlanDeviation} and~\ref{subsection:PredictingUpcomingActions}. Algorithm~\ref{alg:PlanOptimalityMonitoring} formally describes our planning-based approach to detect sub-optimal plan steps.  
The algorithm takes as input a plan optimality monitoring problem $T_{\pi^{*}}$ (Definition~\ref{def:OptimalityMonitoring}), i.e., a planning domain, an initial state, a monitored goal, and a set of observed actions as the execution of an agent plan. 
The algorithm initially computes key information using the landmark extraction algorithm proposed by Hoffman~et al.~\cite{Hoffmann2004_OrderedLandmarks} (using the function \textsc{ExtractLandmarks}). 
Afterwards, it analyses plan execution by iterating over the set of observed actions and applying them, checking which actions do not contribute to the monitored goal. Any such action that does not contribute to achieve the monitored goal is then considered to be sub-optimal. When analysing plan execution deviation (via the distance to the monitored goal) our algorithm can use any domain-independent heuristic, and to do so, we estimate goal distance using the function \textsc{EstimateGoalDistance}\footnote{In Section~\ref{subsection:Heuristics}, we provide a list of domain-independent heuristics that we used in our experiments to estimate goal distance.}. We use Algorithm~\ref{alg:compute} (\textsc{PredictUpcomingActions}) to predict upcoming actions via landmark consideration, in turn utilising \textsc{Max-Heuristic} due to its admissibility, and the fact that it estimates costs for only a short distance (0 or 1). 
The ``if'' statement on Line~\ref{alg:Monitor:Estimatesub-optimalstep} combines heuristic cost estimation and landmark based action prediction, labelling a step as sub-optimal if an observed action is not in the set of predicted upcoming actions, and the estimated distance of the current state is greater than the previous one. 

\floatname{algorithm}{Algorithm}
\begin{algorithm}[h!]
    \caption{Plan Optimality Monitoring.}
	\begin{flushleft}
    \textbf{Parameters:} $\Xi$ $=$ $\langle$$\Sigma$, $\mathcal{A}$$\rangle$ \textit{planning domain}, $\mathcal{I}$ \textit{initial state}, $G$ \textit{monitored goal}, and $O$ \textit{observed actions}.
    \\\textbf{Output:} $A_{sub-optimal}$ \textit{as sub-optimal actions}.
	\end{flushleft}
	\label{alg:PlanOptimalityMonitoring}
    \begin{algorithmic}[1]
        \Function{MonitorPlanOptimality}{$\Xi$,$\mathcal{I}$,$G$,$O$}
        		\State $A_{sub-optimal} \gets \emptyset$ \Comment{\textit{Actions that do not contribute to achieve the monitored goal $G$.}}
        		\State $\mathcal{L} \gets $ \textsc{ExtractLandmarks($\mathcal{I}$, $G$)}
			\State $\state \gets \mathcal{I}$ \Comment{\textit{$\state$ is the current state.}}
			\State $\eta_{PActions} \gets $ \textsc{PredictUpcomingActions}($\Xi$, $\state$, $\mathcal{L}$)
			\State $D_{G} \gets$ \textsc{EstimateGoalDistance}($\state$, $G$) \Comment{\textit{A domain-independent heuristic to estimate goal $G$ from $\state$.}}
        		\For{each observed action $o$ in $O$}
        			\State $\state \gets \state$.\textsc{Apply}($o$)
        			\State $D'_{G} \gets$ \textsc{EstimateGoalDistance}($\state$, $G$)
        			\If{$o$ $\notin$ $\eta_{PActions}$ $\wedge$ $(D'_{G} > D_{G})$}\label{alg:Monitor:Estimatesub-optimalstep}
              		\State $A_{sub-optimal} \gets A_{sub-optimal} \cup o$
            		\EndIf
				\State $\eta_{PActions} \gets$ {\textsc{PredictUpcomingActions}($\Xi$, $\state$, $\mathcal{L}$)}
        			\State $D_{G} \gets D'_{G}$
        		\EndFor
        		\State \textbf{return} $A_{sub-optimal}$
        \EndFunction
    \end{algorithmic}
\end{algorithm}

We note that our approach iterates over the observation sequence $O$ after extracting landmarks  for $G$, and during each iteration, and also iterates over all fact landmarks $\mathcal{L}$ to predict non-regressive actions, and after, it calls a heuristic function  to estimate the distance to $G$. If the complexity of extracting landmarks is $\mathit{EL}$, and of running the heuristic function $\mathit{HF}$, then the complexity of our plan optimality monitoring approach is bounded by $O(\mathit{EL} + |O| \cdot |\mathcal{L}| \cdot \mathit{HF})$.

\section{Detecting Commitment Abandonment}
\label{section:CommitmentAbandonment}

In this section, we apply our approach for plan optimality monitoring to infer when agents are likely to abandon commitments to each other. 
Consider the following scenario. 
An agent $A_1$ delegates a goal $G_1$ to an agent $A_2$ such that $A_2$ is now committed to achieving $G_1$, and consider that $G_1$ can be achieved optimally through a plan consisting of actions $o^{G1}_1, o^{G1}_2, o^{G1}_3$. 
However, instead of executing any of these actions, $A_{2}$ executes $o^{?}_1, o^{?}_2, o^{G1}_1, o^{?}_3$. 
In this situation, agent $A_1$ needs to determine whether or not $A_2$ is still committed to $G_1$ or not, and decide whether to assume $A_2$ abandoned $G_1$, and delegate $G_1$ to another agent. 
Motivated by such scenario, we formally define the commitment abandonment problem and then develop an approach to efficiently solve this problem using fact partitions (Section~\ref{sec:fact-partitioning}) and the techniques from Section~\ref{section:PlanOptimalityMonitoring}. 

\subsection{Commitment Abandonment Problem}
\label{subsection:CommitmentAbandonmentProblem}

We define \emph{commitment abandonment} as a situation in which an agent switches from executing the actions of one plan that achieves the consequent it is committed to, by executing actions from another plan. 
This plan may achieve other goals, including the consequent of other commitments, or the agent has no intention to achieve its original commitment.
Actions in a plan that do not contribute to achieve the consequent of a commitment may indicate that the debtor agent is likely to abandon this commitment.
An agent may abandon a commitment for a variety of reasons. For example, it may have conflicting commitments, and must abandon the less important commitment to achieve the more important one.



Here, we take inspiration from earlier work \cite{Commitments_MeneguzziTS13,CommitmentsJaCaMo_BaldoniBCM15} that connects commitments to planning, so the domain definition $\Xi$ represents the environment where agents can interact and act, i.e., $\Sigma$ is set of environment properties and $\mathcal{A}$ is a set of available actions. 
Now, consider a commitment \texttt{C(DEBTOR, CREDITOR, At, Ct)}: for a \texttt{DEBTOR} to achieve the consequent \texttt{Ct} from the antecedent \texttt{At} we require that: (i) the antecedent \texttt{At} must be in the initial state $\mathcal{I}$, i.e., \texttt{At} $\subseteq \mathcal{I}$; and (ii) the consequent \texttt{Ct} is the goal $G$. 
Thus, a plan $\pi$ for \texttt{C(DEBTOR, CREDITOR, At, Ct)} is a sequence of actions $[a_1, a_2, ..., a_n]$ (where $a_i \in \mathcal{A}$) that modifies the state \texttt{At} $\subseteq \mathcal{I}$ into one where \texttt{Ct} holds by the successive (ordered) execution of actions in a plan $\pi$. We note that both antecedent \texttt{At} and consequent \texttt{Ct} consist of a set of facts, more specifically, both are states.

To decide if a debtor will abandon a commitment, we monitor its behaviour in an environment by observing its successive execution of actions. 
This successive execution of actions represents an observation sequence (Definition~\ref{def:Observation}) that should achieve a consequent from an antecedent. 
When a \texttt{DEBTOR} commits to an agent \texttt{CREDITOR} to bring about the \texttt{consequent} of a commitment, the \texttt{DEBTOR} should individually commit to achieving such a \texttt{consequent} state,
and to achieve such state, the \texttt{DEBTOR} has to execute a plan. 
An observer does not have access to \texttt{DEBTOR}'s internal state, and consequently to what plan it has committed to. Therefore, when there are multiple optimal plans, we need to be able to determine which of those plans the \texttt{DEBTOR} is pursuing. 
Thus, in Definition~\ref{def:IndividualCommitment}, we formally define an individual commitment from an observer's point of view. 

\begin{definition} [\textbf{Individual Commitment}] \label{def:IndividualCommitment}
Given a set of plans, a \texttt{DEBTOR} agent is individually committed to a plan $\pi$ if, given a sequence of observations $o_1, \ldots, o_m$: i) $o_k \in \pi$ where $(1 \leq k\leq m)$; and ii) if $o_k=a_j$, then $\forall i=1 \ldots j-1$, $a_i \in O$ and $a_i$ occurs before $a_{i+1}$ in $O$. 
An observation $o_p$ does not contribute to achieve a consequent \texttt{Ct} if the \texttt{DEBTOR} agent is committed to plan $\pi$ and: $o_p \notin \pi$; or if $o_p=a_j$, $a_{k}$ has not yet been observed where $k<j$. 
\end{definition}

Finally, using the notion of an individual commitment, we formally define a commitment abandonment problem over a planning theory in terms of a large enough deviation from such an individual commitment in Definition~\ref{def:CommitmentAbandonment}. 
Note that with Definition~\ref{def:IndividualCommitment}, we can now think of deviations from observations that constitute a strict sub-sequence of any optimal plan that start with the initial state, allowing an agent to infer abandonment at any point in a partial plan execution.  

\begin{definition} [\textbf{Commitment Abandonment Problem}] \label{def:CommitmentAbandonment}
A commitment abandonment problem is encoded as  a tuple $CA = $ $\langle \Xi, C, \mathcal{I}, O, \theta \rangle$, in which $\Xi$ is a planning domain definition; 
$C$ is the commitment, in which C(\texttt{DEBTOR}, \texttt{CREDITOR}, At, Ct), \texttt{DEBTOR} is the debtor, \texttt{CREDITOR} is the creditor, At is the antecedent condition, and Ct is the consequent. $\mathcal{I}$ is the initial state (s.t., At $\subseteq \mathcal{I}$); $O$ is an observation sequence of the plan execution with each observation $o_i \in O$ being an action from domain definition $\Xi$;
and $\theta$ is a threshold that represents the permitted fraction of actions (in relation to an individually committed from Definition~\ref{def:IndividualCommitment}) in an observation sequence that do not contribute to achieving Ct from At $\subseteq \mathcal{I}$ in which the \texttt{DEBTOR} can execute in $O$.
\end{definition}


The solution for a commitment abandonment problem is whether an observation sequence $O$ has deviated more than $\theta$ from the optimal plan to achieve the consequent $Ct$ of commitment $ C$. In this work, we use the threshold $\theta$ as a degree of tolerance to sub-optimality that the creditor affords the debtor.

\subsection{Detecting Commitment Abandonment via Optimality Monitoring}
\label{subsection:DetectingCommitmentAbandonment}

To detect commitment abandonment, we infer sub-optimal action steps combining the techniques from Section~\ref{section:PlanOptimalityMonitoring} and use the concept of fact partitions from Section~\ref{sec:fact-partitioning}. 
Once we observe evidence of such fact partitions in the observations we can determine that a goal is no longer achievable. 
We extract fact partitions using a function called \textsc{PartitionFacts}.


Algorithm~\ref{alg:DetectCommitmentAbandonment} formalises our approach to solve a commitment abandonment problem.
The algorithm takes as input a commitment abandonment problem $\textit{CA}$ and returns whether a commitment has been abandoned, based on whether one of the following occurs during plan execution: (1) if \textit{Strictly Activating} facts that we extracted are not in the initial state (Line \ref{alg:DetectAbandonment:SA}); 
(2) if we observe the evidence of any \textit{Unstable Activating} and \textit{Strictly Terminal} facts during the execution of actions in the observations (Line \ref{alg:DetectAbandonment:STandUA}); 
or (3) if the number of sub-optimal action steps are greater than the threshold $\theta$ (i.e., the percentage of actions away from optimal execution that the creditor allows the debtor to deviate in achieving the consequent state) defined by the creditor (Line \ref{alg:DetectAbandonment:sub-optimalAndThreshold}). 
If none of these conditions hold, the debtor is considered to remain committed to achieving the consequent state of the commitment.
Note that in condition (2), the presence of predicates from two of the fact partitions can determine that the monitored goal (or consequent) is unreachable, because there is no available action that can make the facts hold. 

The checks from condition (3) is substantially different than the other conditions, since this is a \emph{subjective} measure of abandonment, which we capture in the threshold $\theta$. 
We need to use this subjective measure because in most realistic domains, there is never a definite logical condition (which is what we capture with the Fact Partitions) that tells that the commitment is irreparably abandoned, and this threshold allows us to reason about such non-clear-cut situations. 
One can think about the threshold as a subjective measure of \emph{patience} on the part of the creditor of a commitment. 
Specifically, this captures the amount of slack given to a debtor when the debtor interleaves its other goals with the delegated one. 

\floatname{algorithm}{Algorithm}
\begin{algorithm}[hb!]
    \caption{Detecting Commitment Abandonment via Plan Optimality Monitoring and Fact Partitions.}
	\begin{flushleft}
    \textbf{Parameters:} $\Xi = \langle \Sigma$, $\mathcal{A} \rangle$ \textit{planning domain}, $At$ \textit{antecedent condition} ($At\subseteq \mathcal{I}$), $Ct$ \textit{consequent condition}, $\mathcal{I}$ \textit{initial state}, $O$ \textit{observation sequence}, and $\theta$ \textit{threshold}.
    \\\textbf{Output:} True or False.
	\end{flushleft}
    \label{alg:DetectCommitmentAbandonment}
    \begin{algorithmic}[1]
        \Function{HasAbandoned}{$\Xi$, $At$, $Ct$, $\mathcal{I}$, $O$, $\theta$}
        	\State $\langle F_{sa}, F_{ua}, F_{st} \rangle \gets \Call{PartitionFacts}{\Sigma, \mathcal{A}}$ \label{alg:DetectAbandonment:factPartitioner}
			\If{$F_{sa} \cap (At \subseteq \mathcal{I}) = \emptyset$}\label{alg:DetectAbandonment:SA}
				\State \textbf{return true} \Comment{\textit{Ct is no longer possible.}}
			\EndIf            
            \For{each observed action $o$ in $O$}
        		\State $\state \gets \state$.\textsc{Apply}($o$)
            	\If{$(F_{ua} \cup  F_{st}) \subseteq \state$} \label{alg:DetectAbandonment:STandUA}
					\State \textbf{return true} \Comment{\textit{Ct is no longer possible.}}
				\EndIf
			\EndFor
            \State $A_{sub-optimal} \gets$ \Call{MonitorPlanOptimality}{$\Xi$, $\mathcal{I}$, $Ct$, $O$} \label{alg:DetectAbandonment:sub-optimalAndThreshold}
            \If{$|A_{sub-optimal}| > (\theta * |O|)$}
            	\State \textbf{return true} \Comment{\textit{Debtor has abandoned the commitment.}}
            \EndIf
            \State \textbf{return false} \Comment{\textit{Debtor may still be committed to achieve Ct.}} \label{alg:DetectAbandonment:debtorCommitted}
        \EndFunction
    \end{algorithmic}
\end{algorithm}

\subsection{Working Example}
\label{subsection:WorkingExample}

\begin{figure}[b!]
  \centering
  \includegraphics[width=0.85\linewidth]{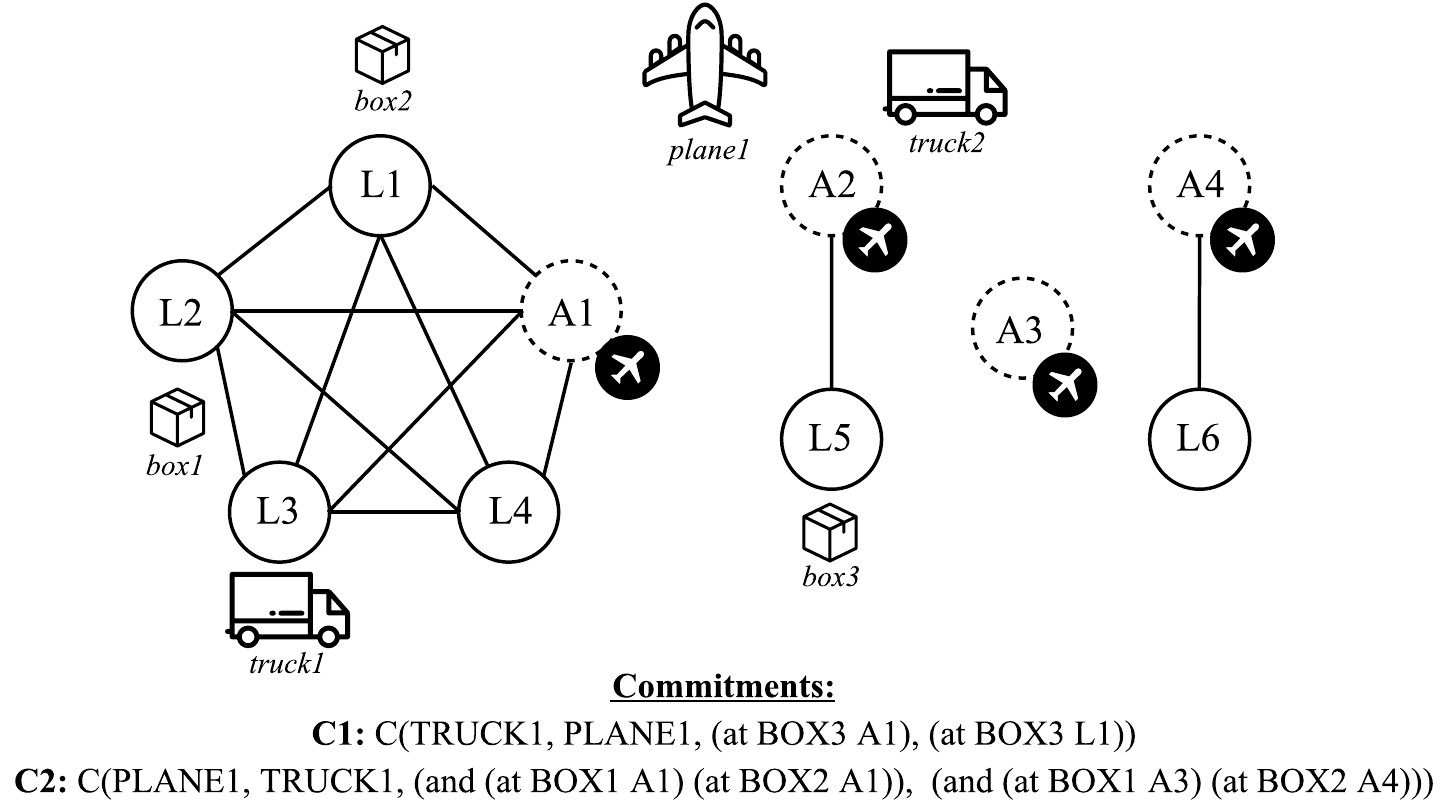}
  \caption{\textsc{Logistics} working example.}
  \label{fig:LogisticsWorkingExample}
\end{figure}

To exemplify how our approaches detect sub-optimal action steps and determine commitment abandonment, consider the \textsc{Logistics} problem example shown in Figure~\ref{fig:LogisticsWorkingExample}. 
This example formalises two commitments: C1 represents that the debtor agent \texttt{TRUCK1} is committed to the creditor agent \texttt{PLANE1} to bring about the consequent \texttt{(at BOX3 L1)} when the antecedent \texttt{(at BOX3 A1)} becomes true; and C2 represents that the debtor agent \texttt{PLANE1} is committed to the creditor agent \texttt{TRUCK1} to bring about the consequent \texttt{(and (at BOX1 A3) (at BOX2 A4))} when the antecedent \texttt{(and (at BOX1 A1) (at BOX2 A1))} becomes true. 
Assuming that for C1 the threshold $\theta$ is 0, and for C2 the threshold $\theta$ is 0.3, Tables~\ref{tab:workingexample1} and~\ref{tab:workingexample2} show observed actions for C1 and C2, respectively. 
Rows in grey represent sub-optimal actions, and rows without a number (i.e., -) represent actions executed by the creditor agent that are going to achieve the antecedent state.

From the observation sequence shown in Table~\ref{tab:workingexample1} and the threshold $\theta$ = 0\%, our approach returns that the observations at time 1 and 2 are sub-optimal actions, and therefore debtor agent \texttt{TRUCK1} has abandoned commitment C1, since $\theta$ = 0 (i.e., the creditor does not allow any deviation), and the agent has executed two actions that do not contribute to achieve the consequent state of C1. 
The observed action at time 3 is an optimal action because the agent is moving towards the location L1, where it must unload \texttt{BOX3}.

Now consider the observation sequence in Table~\ref{tab:workingexample2} and a threshold $\theta$ = 0.3. While our approach returns that the observations at time 3 and 4 are sub-optimal actions, the threshold (which allows $9 \times 0.3=2.7$ sub-optimal actions)  means that the debtor agent \texttt{PLANE1} is considered to remain committed to achieve the consequent of C2.

\begin{table}[hb!]
	\footnotesize
	\begin{center}
	\begin{tabular}{ll}
        - & \texttt{(loadAirplane BOX3 PLANE1 A2)} \\
        - & \texttt{(fly PLANE1 A2 A1)} \\
        - & \texttt{(unloadAirplane BOX3 PLANE1 A1)} \\
        0 & \texttt{(loadTruck BOX3 TRUCK1 A1)} \\
        \cellcolor[HTML]{9B9B9B}{\color[HTML]{333333} 1} & \cellcolor[HTML]{9B9B9B}{\color[HTML]{333333} \texttt{(drive TRUCK1 A1 L4 CITY1)}} \\
        \cellcolor[HTML]{9B9B9B}{\color[HTML]{333333} 2} & \cellcolor[HTML]{9B9B9B}{\color[HTML]{333333} \texttt{(drive TRUCK1 L4 L2 CITY1)}} \\
        3 & \texttt{(drive TRUCK1 L2 L1 CITY1)} \\
	\end{tabular}
	\end{center}
	\caption{Observation sequence (1), monitoring \texttt{TRUCK1}.}
	\label{tab:workingexample1}
\end{table}

\begin{table}[ht!]
	\footnotesize
	\begin{center}
	\begin{tabular}{ll}
        - & \texttt{(drive TRUCK1 L2 A1 CITY1)} \\
        - & \texttt{(unloadTruck BOX2 TRUCK1 A1)} \\
        - & \texttt{(unloadTruck BOX1 TRUCK1 A1)} \\
        0 & \texttt{(fly PLANE1 A2 A1)} \\
        1 & \texttt{(loadAirplane BOX2 PLANE1 A1)} \\
        2 & \texttt{(loadAirplane BOX1 PLANE1 A1)} \\
		\cellcolor[HTML]{9B9B9B}{\color[HTML]{333333} 3} & \cellcolor[HTML]{9B9B9B}{\color[HTML]{333333} \texttt{(fly PLANE1 A1 A2)}} \\
		\cellcolor[HTML]{9B9B9B}{\color[HTML]{333333} 4} & \cellcolor[HTML]{9B9B9B}{\color[HTML]{333333} \texttt{(fly PLANE1 A2 A1)}} \\
        5 & \texttt{(fly PLANE1 A1 A3)} \\
        6 & \texttt{(unloadAirplane BOX1 PLANE1 A3)} \\
        7 & \texttt{(fly PLANE1 A3 A4)} \\
        8 & \texttt{(unloadAirplane BOX2 PLANE1 A4)} \\
	\end{tabular}
	\end{center}
	\caption{Observation sequence (2), monitoring \texttt{PLANE1}.}
	\label{tab:workingexample2}
\end{table}

\section{Experiments and Evaluation}
\label{section:ExperimentsEvaluation}

In this section, we describe the experiments and evaluation we carried out on our approaches. 
In Section~\ref{subsection:DomainsDatasets}, we describe the planning domains and the datasets we used as a benchmark to our proposed approaches. 
In Section~\ref{subsection:Metrics}, we show the set of metrics we used for evaluation. In Section~\ref{subsection:Heuristics}, we describe the domain-independent heuristics we used in our experiments.
Finally, in Sections~\ref{subsection:OptimalityMonitoringExperiments} and~\ref{subsection:CommitmentAbandonmentExperiments}, we show the experiments and evaluation on our plan optimality monitoring approach and our commitment abandonment approach. To evaluate our approaches, we ran all experiments using a single core of a 12 core Intel(R) Xeon(R) CPU E5-2620 v3 @ 2.40GHz with 16GB of RAM. 
The Java virtual machine we ran the experiments on was limited to 1GB of memory, and we imposed a  one minute time-out for all our experiments. 

\subsection{Domains and Datasets}
\label{subsection:DomainsDatasets}

We empirically evaluated our approaches (plan optimality monitoring and commitment abandonment detection) over several widely used planning domains, most of which are inspired by real-world scenarios. 
Most domains we used are from the Artificial Intelligence Planning and Scheduling (AIPS) competitions in 1998, 2000, and 2002~\cite{AIPS_98,AIPS_2000,AIPS_2002}, and goal and plan recognition datasets~\cite{Pereira_Meneguzzi_PRDatasets_2017}.
The \textsc{Driver-Log} domain consists of drivers that can walk between locations and trucks that can drive between locations, and goals consist of transporting packages between locations. 
\textsc{Depots} combines transportation and stacking, in which goals involve moving and stacking packages by using trucks and hoists between depots. 
\textsc{Easy-IPC-Grid} consists of an agent that moves in a grid from cells to others by transporting keys to open locked locations for releasing agents that are at isolated cells. 
The \textsc{Ferry} domain consists of set of cars that must be moved to desired locations using a ferry that can carry only one car at a time. 
\textsc{Logistics}, described previously, consists of airplanes and trucks transporting packages between locations (e.g., airports and cities). 
\textsc{Satellite} involves using one or more satellites to make observations by collecting data and downloading the data to a desired ground station. 
\textsc{Sokoban} involves pushing a set of boxes into specified locations in a grid with walls. 
Finally, \textsc{Zeno-Travel} is a domain where passengers can embark and disembark onto aircraft that can fly at two alternative speeds between locations. We select and use these domains in our datasets because they are inspired by real-world scenarios, and most of them contain and deal with more than one agent and several objects in the environment.

For each of these domains we selected 15 to 30 non-trivial problem instances\footnote{A non-trivial planning problem contains a large search space (in terms of search branching factor and depth), and therefore, even modern planners such as \textsc{Fast-Downward (FD)} takes up to a 5-minute to complete. 
In our datasets, the number of instantiated (grounded) actions is between 146 and 4322, and plan length is between 12.2 and 25.7.}, with each problem instance also associated to a set of observations (i.e., plan executions), i.e., a plan optimality monitoring problem. 
This set of observations can represent either an optimal or a sub-optimal plan execution. 
We generate plans (optimal and sub-optimal) using open-source planners, such as \textsc{BlackBox}, \textsc{Fast-Downward}, \textsc{FF}, and \textsc{LAMA}~\cite{RichterLPG_2010}. 
For sub-optimal plans, we (manually) annotated the sub-optimal action steps and how many sub-optimal steps each plan has. 
These steps consist of actions that do not contribute to achieving the monitored goal, representing steps that our plan optimality monitoring approach aims to detect. 
We manually annotated the sub-optimal steps in the sub-optimal plans in order to define the exact time steps in which the sub-optimal actions happened during a plan execution. 
The key goals in optimality monitoring is to detect not only the \textit{number} of sub-optimal steps (a relatively easy problem) in a plan execution but also the \textit{exact time steps} in which such sub-optimal actions happened in a plan. 

We built a dataset for the experiments on detecting commitment abandonment using 30 commitment abandonment problems (10 problems for each threshold value: 0\%, 10\%, and 30\%), and for these problems we generated plans (observed actions) that either abandoned (ultimately went to a different goal or consequent) or did not abandon their corresponding goals/consequents, varying the number of abandoned actions. 
For instance, there are commitment abandonment problems that contain plans with sub-optimal steps that do not abandon the defined goals/consequents, and it happens due to the fact that the number of sub-optimal steps is not greater than the permitted fractions of actions ($\theta$) that are allowed to deviate during the plan execution.
Like the dataset for plan optimality monitoring, we use non-trivial problem instances to define the commitments in our dataset for commitment abandonment detection, varying the number of observations (i.e., plan length) between 10.0 and 23.5 ($|O|$). 
Our dataset ensures that the extra actions added to all sub-optimal plans do not make the actual monitored goal unreachable or invalidate the plan. 


\subsection{Evaluation Metrics}
\label{subsection:Metrics}


Using the generated datasets we evaluated how accurately our approaches detects actions that do not contribute to achieve a corresponding monitored goal and commitment abandonment by using the following metrics. 
\textit{Precision} (Positive Predictive Value - PPV) is the ratio between true positive results, and the sum of true positive and false positive results. \textit{Precision} provides the percentage of positive predictions that is correct. \textit{Recall} (True Positive Rate - TPV) is the ratio between true positive results, and the sum of the number of true positives and false negatives. \textit{Recall} provides the percentage of positive cases that our approaches have detected. 
The \textit{F1-score} (F1) is a measure of accuracy that aims to provide a trade-off between \textit{Precision} and \textit{Recall}.

\subsection{Planning Heuristics}
\label{subsection:Heuristics}

Since our approaches can exploit any domain-independent heuristic to compute whether an action contributes to goal achievement, we evaluated our approaches using several admissible and inadmissible heuristics from the planning literature, as follows. 



\begin{itemize}\label{experiment:heuristics}
	\item \textsc{Max-Heuristic} ($h_{\mathit{max}}$) is an admissible heuristic proposed by Bonet and Geffner \cite{BonetGeffner_PlanningAsHeuristic_2001},  
    and this heuristic is based on the delete-list relaxation, in which delete-effects of actions are ignored during calculation of the heuristic cost to a goal. This calculation is the cost of a conjunctive goal, which represents the maximum cost to achieve each of the individual facts;
	\item \textsc{Sum} ($h_{\mathit{sum}}$) is also an admissible heuristic proposed by Bonet and Geffner \cite{BonetGeffner_PlanningAsHeuristic_2001}, 
    and this heuristic works similarly to \textsc{Max-Heuristic}. However, the \textsc{Sum} heuristic is often more informative than \textsc{Max-Heuristic};
	\item \textsc{Adjusted-Sum} ($h_{\mathit{adjsum}}$)~\cite{Heuristics_NguyenAndKambhampati_2002} is an inadmissible heuristic that improves the \textsc{Sum} heuristic by taking into account both negative and positive interactions among facts;
	\item \textsc{Adjusted-Sum2} ($h_{\mathit{adjsum2}}$)~\cite{Heuristics_NguyenAndKambhampati_2002} is an inadmissible heuristic that improves its previous version (\textsc{Adjusted-Sum}) by combining the computation of the \textsc{Set-Level} heuristic\footnote{The \textsc{Set-Level} heuristic estimates the cost to a goal by returning the level of the planning graph where all facts of the goal sate are reached without any mutex free~\cite{Heuristic_Nguyen_2000}.} and the relaxed plan heuristic;
	\item \textsc{Adjusted-Sum2M} ($h_{\mathit{adjsum2M}}$)~\cite{Heuristics_NguyenAndKambhampati_2002} is an inadmissible heuristic that improves the \textsc{Adjusted-Sum2};
	\item \textsc{Combo} ($h_{\mathit{combo}}$)~\cite{Heuristics_NguyenAndKambhampati_2002} is an inadmissible heuristic that improves the \textsc{Adjusted-Sum} by combining the computation of the \textsc{Adjusted-Sum} heuristic and the \textsc{Set-Level} heuristic; and
	\item \textsc{Fast-Forward} ($h_{\mathit{ff}}$) is a well-known inadmissible heuristic in the planning community~\cite{FFHoffmann_2001} that relies on state-space search and estimates the goal distance by using the delete-list relaxation.
\end{itemize}


\subsection{Plan Optimality Monitoring Experiments}
\label{subsection:OptimalityMonitoringExperiments}


For experiments and evaluation on our plan optimality monitoring approach, we use the metrics presented before, as follows: \textit{Precision} (PPV), \textit{Recall} (TPR), and \textit{F1-score} (F1). Here, for detecting sub-optimal action steps, true positive results represent the number of sub-optimal actions detected that do not contribute to achieve the monitored goal. False positive results represent the number of actions that our approach labelled as a sub-optimal action which is in fact an optimal action. False negative is a sub-optimal action that is not detected by our approach. 

\frm{I don't like the way this paragraph was written, but I don't have much inspiration to rewrite it right now.}
\rfp{Let's talk about this. Here, I just tried to explain the metrics we used to evaluate our plan optimality monitoring approach, i.e., by explaining what is a TP, FP, and FN in the context of optimality monitoring. Do you understand now?! I think it is convenient to make it clear for the reviewers, and I did that before presenting the experiments.}

We separated our experiments in three different parts, evaluating our plan optimality monitoring techniques separately (techniques from Sections~\ref{subsection:AnalysingPlanDeviation}~and~\ref{subsection:PredictingUpcomingActions}) and together (approach from Section~\ref{subsection:Detectingsub-optimalSteps}), as follows: (1) Table~\ref{tab:Table1:OptimalityMonitoring-Landmarks} shows experimental results by using just our technique for predicting upcoming actions via landmarks for plan optimality monitoring; (2) Tables~\ref{tab:Table1:OptimalityMonitoring-PlanDeviation}~and~\ref{tab:Table2:OptimalityMonitoring-PlanDeviation} show experimental results by using just our technique for analysing plan execution deviation using all domain-independent heuristics we presented before; and finally, (3) Tables~\ref{tab:Table1:OptimalityMonitoring} and~\ref{tab:Table2:OptimalityMonitoring} show experimental results of our plan optimality monitoring approach combining the prediction of upcoming actions and the plan deviation detection approach (as we developed in Section~\ref{subsection:Detectingsub-optimalSteps}). 
Each row contains results for all domains averaged over all problem instances for the set of observations $|O|$ (i.e., the number of actions in a plan execution --- plan length), the average number of extracted fact landmarks $\mathcal{L}$, monitoring time (in seconds), and metrics (\textit{Precision} -- PPV, \textit{Recall} -- TPR, and the \textit{F1-score} -- F1).  
The observation averages of column $|O|$ range between 12.2 and 25.7, indicating that all plans we analyse are non-trivial plan executions. 

\begin{table*}[ht!]
\centering
\begin{tabular}{lllllll}
\hline
		\multicolumn{7}{c}{\begin{tabular}[c]{@{}c@{}} Plan Optimality Monitoring \\ (Predicting Upcoming Actions via Landmarks) \end{tabular}}	\\ \hline
		\multicolumn{1}{l}{\textbf{Domain}} & \multicolumn{1}{l}{\textbf{$|O|$}} & \textbf{$|\mathcal{L}|$} & \textbf{Time}    & \multicolumn{3}{l}{\textbf{PPV / TPR / F1}} \\ \hline
		
\multicolumn{1}{l}{\textsc{Driver-Log} (20)}    & \multicolumn{1}{l}{20.1}         & 53.6         & 0.19             & \multicolumn{3}{l}{18.8\% / 72.5\% / 29.8\%} \\ 

\multicolumn{1}{l}{\textsc{Depots} (30)}        & \multicolumn{1}{l}{16.7}         & 64.7         
& 0.33             & \multicolumn{3}{l}{30.7\% / 65.4\% / 41.8\%} \\ 

\multicolumn{1}{l}{\textsc{Easy-IPC-Grid} (30)} & \multicolumn{1}{l}{14.1}         & 48.5         & 0.26             & \multicolumn{3}{l}{34.9\% / 74\% / 47.4\%} \\ 

\multicolumn{1}{l}{\textsc{Ferry} (30)}         & \multicolumn{1}{l}{13.8}         & 18.1         & 0.06             & \multicolumn{3}{l}{41.4\% / 96.6\% / 58\%}	\\ 

\multicolumn{1}{l}{\textsc{Logistics} (30)}     & \multicolumn{1}{l}{20.8}         & 24.0           & 0.20             & \multicolumn{3}{l}{26.3\% / 86.9\% / 40.4\%} \\ 

\multicolumn{1}{l}{\textsc{Satellite} (20)}     & \multicolumn{1}{l}{25.7}         & 60.8         & 1.10             & \multicolumn{3}{l}{13.1\% / 70.2\% / 22.0\%} \\ 

\multicolumn{1}{l}{\textsc{Sokoban} (20)}       & \multicolumn{1}{l}{24.0}           & 76.5         & 0.49             & \multicolumn{3}{l}{1.4\% / 21.4\% / 2.7\%} \\ 

\multicolumn{1}{l}{\textsc{Zeno-Travel} (15)}   & \multicolumn{1}{l}{12.2}         & 38.7         & 0.21             & \multicolumn{3}{l}{32.7\% / 76.9\% / 45.9\%} \\ \hline
\end{tabular}
\caption{Experimental results for detecting sub-optimal action steps using the technique Predicting Upcoming Actions via Landmarks (1).}
\label{tab:Table1:OptimalityMonitoring-Landmarks}
\end{table*}

Table~\ref{tab:Table1:OptimalityMonitoring-Landmarks} shows the results of our optimality monitoring technique for predicting upcoming landmarks. 
These results show that our technique is quite fast (at most 1 second) and relatively accurate for all domains but \textsc{Sokoban}. 
Apart from \textsc{Sokoban}, the TPR results are higher than 65.4\%. 
For \textsc{Logistics} and \textsc{Ferry} instances this technique yields almost perfect results with respect to TPR, but the results for PPV and F1 are quite low, at less than 50\% on average. 
Other domains result in lower FPR and F1 in detecting sub-optimal action steps in a plan. 
Thus, the technique can be useful on its own in correctly detecting upcoming actions via landmarks, showing good results for TPR, which means that it is accurate to detect the set of actions that dot not contribute to achieve the monitored goal (true positive results). 
However, in general, for all domains, the technique that uses landmarks alone is not precise enough to detect sub-optimal actions, since it returns many false positive actions that do not contribute to achieve the monitored goal.


Tables~\ref{tab:Table1:OptimalityMonitoring-PlanDeviation}~and~\ref{tab:Table2:OptimalityMonitoring-PlanDeviation} show performance results for our optimality monitoring approaches using domain-independent heuristics. 
As can be seen in comparison to Table~\ref{tab:Table1:OptimalityMonitoring-Landmarks}, these results are superior to our prediction of upcoming landmarks technique in all evaluated domains and problems for all metrics. 
Some planning heuristics lead to more accurate and faster predictions than others in some planning domains. 
For instance, the \textsc{Fast-Forward} $h_{\mathit{ff}}$ heuristic has near perfect results for PPV, TPR, and F1 for \textsc{Easy-IPC-Grid}, \textsc{Ferry}, \textsc{Logistics}, and \textsc{Zeno-Travel}, whereas for \textsc{Sokoban} and \textsc{Satellite} the results are poor. 
Since all heuristics are relatively cheap to compute, monitoring time is overall very fast, and is, at most, 3.77 seconds. 
Thus, our heuristic approaches to detect plan deviation are generally better performing than predicting upcoming landmarks. 

Tables~\ref{tab:Table1:OptimalityMonitoring} and~\ref{tab:Table2:OptimalityMonitoring} show the results of our plan optimality monitoring approach combining both previous techniques. 
The resulting approach achieves better results than each technique on its own. 
We can see that for the \textsc{Driver-Log} and \textsc{Easy-IPC-Grid} domains our plan optimality monitoring approach yields perfect results (100\% for all metrics) using different heuristics, such as \textsc{Adjusted-Sum2M} ($h_{\mathit{adjsum2M}}$) and \textsc{Fast-Forward} ($h_{\mathit{ff}}$) respectively, along with our technique that predicts upcoming actions via landmarks. 
Apart from the \textsc{Satellite} domain that under-performs for all metrics, our approach is near-perfect monitoring optimality at low runtime in all other planning domains, yielding very good results with different heuristics. 
We note that the poor results with respect to the \textsc{Satellite} domain are related to estimated distance provided by the heuristics. 
We analysed the output of our approach over the problems instances of the \textsc{Satellite} domain, and we observed that all heuristics (for most problem instances) do not detect when the observed actions deviate to achieve the monitored goal, and such issue happens because the heuristics are inaccurate for the problem instances of this particular domain. 
To overcome this issue, we intend to use more modern domain-independent planning heuristics, and then evaluate our approach not only over this domain but also in all domains we used in our experiments.
We also note that some heuristics outperform others for the same domain, for instance, the \textsc{Adjusted-Sum2M} heuristic under-performs against others for the \textsc{Ferry} domain. 
In summary, our approach yields very good results detecting sub-optimal plan steps in deterministic planning domains, with most domains having high \textit{F1-scores} from perfect and near-perfect accuracy (depending on the heuristic), while only one domain (i.e., \textsc{Satellite}) has relatively inaccurate results. We also note that the combination of our techniques yield better results than using the techniques separately, over-performing for all metrics, as we show in Tables~\ref{tab:Table1:OptimalityMonitoring-Landmarks},~\ref{tab:Table1:OptimalityMonitoring-PlanDeviation}, and~\ref{tab:Table2:OptimalityMonitoring-PlanDeviation}.

\afterpage{
\begin{landscape}
\begin{table*}[ht!]
\centering
\begin{tabular}{lllllllllll}
\hline
		\multicolumn{11}{c}{\begin{tabular}[c]{@{}c@{}} Plan Optimality Monitoring \\ (Analysing Plan Execution Deviation) \end{tabular}}	\\ \hline
		&                                   &              & \multicolumn{4}{c}{$h_{\mathit{adjsum}}$}                                     & \multicolumn{2}{c}{$h_{\mathit{adjsum2}}$}            & \multicolumn{2}{c}{$h_{\mathit{adjsum2M}}$}           \\ \hline
\multicolumn{1}{l}{\textbf{Domain}}    & \multicolumn{1}{l}{\textbf{$|O|$}} & \textbf{$|\mathcal{L}|$} & \textbf{Time}    & \multicolumn{3}{l}{\textbf{PPV / TPR / F1}} & \textbf{Time} & \textbf{PPV / TPR / F1} & \textbf{Time} & \textbf{PPV / TPR / F1} \\ \hline

\multicolumn{1}{l}{\textsc{Driver-Log} (20)}    & \multicolumn{1}{l}{20.1}         & 53.6         
{\hskip 12pt} & 0.15			& \multicolumn{3}{l}{63.6\% / 77.8\% / 70\%}
{\hskip 12pt} & 0.23			& 47.2\% / 94.4\% / 63\%       
{\hskip 12pt} & 0.65			& 47.4\% / 100\% / 64.3\%		\\

\multicolumn{1}{l}{\textsc{Depots} (30)}        & \multicolumn{1}{l}{16.7}         & 64.7         
& 0.27			& \multicolumn{3}{l}{46\% / 88.5\% /60.5\%}
& 0.39			& 47.2\% / 96.2\% / 63.3\%
& 1.01			& 47.4\% / 96.4\% / 63.5\%      \\

\multicolumn{1}{l}{\textsc{Easy-IPC-Grid} (30)} & \multicolumn{1}{l}{14.1}         & 48.5         
& 0.28			& \multicolumn{3}{l}{83.9\% / 92.9\% / 88.1\%}
& 0.37			& 100\% / 96.2\% / 98\%
& 0.46			& 86.4\% / 79.2\% / 82.6\%        \\

\multicolumn{1}{l}{\textsc{Ferry} (30)}         & \multicolumn{1}{l}{13.8}         & 18.1         
& 0.13			& \multicolumn{3}{l}{57.6\% / 67.9\% / 62.3\%}
& 0.17			& 100\% / 78.6\% / 88\%
& 0.34			& 75.0\% / 42.9\% / 54.5\%        \\

\multicolumn{1}{l}{\textsc{Logistics} (30)}     & \multicolumn{1}{l}{20.8}         & 24.0           
& 0.13			& \multicolumn{3}{l}{100\% / 13\% / 23.1\%}
& 0.17			& 100\% / 91.3\% / 95.5\%       
& 0.59			& 100\% / 91.3\% / 95.5\%       \\

\multicolumn{1}{l}{\textsc{Satellite} (20)}     & \multicolumn{1}{l}{25.7}         & 60.8         
& 0.45			& \multicolumn{3}{l}{100\% / 26.7\% / 42.1\%}
& 0.53			& 45.5\% / 66.7\% / 54.1\%
& 3.21			& 44\% / 73.3\% / 55\%      \\

\multicolumn{1}{l}{\textsc{Sokoban} (20)}       & \multicolumn{1}{l}{24}           & 76.5         
& 1.18			& \multicolumn{3}{l}{35.5\% / 78.6\% / 48.9\%}            
& 1.41			& 5.4\% / 14.3\% / 7.8\%
& 3.50			& 35.5\% / 78.6\% / 48.9\%     \\

\multicolumn{1}{l}{\textsc{Zeno-Travel} (15)}   & \multicolumn{1}{l}{12.2}         & 38.7         
& 0.14			& \multicolumn{3}{l}{77.8\% / 50.0\% / 60.9\%}
& 0.17			& 82.4\% / 100\% / 90.3\%       
& 0.77			& 72.2\% / 92.9\% / 81.3\%       \\ \hline
\end{tabular}
\caption{Experimental results for detecting sub-optimal action steps using the technique Analysing Plan Execution Deviation (1).}
\label{tab:Table1:OptimalityMonitoring-PlanDeviation}
\end{table*}

\begin{table*}[ht!]
\centering
\begin{tabular}{lllllllllll}
\hline
		\multicolumn{11}{c}{\begin{tabular}[c]{@{}c@{}} Plan Optimality Monitoring \\ (Analysing Plan Execution Deviation) \end{tabular}}	\\ \hline
		&                                   & \multicolumn{1}{c}{} & \multicolumn{4}{c}{$h_{\mathit{combo}}$}                                         & \multicolumn{2}{c}{$h_{\mathit{ff}}$}             & \multicolumn{2}{c}{$h_{\mathit{sum}}$}                      \\ \hline
\multicolumn{1}{l}{\textbf{Domain}}    & \multicolumn{1}{l}{\textbf{$|O|$}} & \textbf{$|\mathcal{L}|$}          & \textbf{Time}  & \multicolumn{3}{l}{\textbf{PPV / TPR / F1}} & \textbf{Time} & \textbf{PPV / TPR / F1} & \textbf{Time} & \textbf{PPV / TPR / F1} \\ \hline
\multicolumn{1}{l}{\textsc{Driver-Log} (20)}    & \multicolumn{1}{l}{20.1}         & 53.6         
{\hskip 12pt} & 0.69			& \multicolumn{3}{l}{63.6\% / 77.8\% / 70.0\%}
{\hskip 12pt} & 0.14			& 47.2\% / 94.4\% / 63\%       
{\hskip 12pt} & 0.16			& 63.6\% / 77.8\% / 70\%         \\

\multicolumn{1}{l}{\textsc{Depots} (30)}        & \multicolumn{1}{l}{16.7}         & 64.7         
& 1.03			& \multicolumn{3}{l}{43.6\% / 92.3\% / 59.3\%}
& 0.27			& 47.2\% / 96.2\% / 63.3\%
& 0.29			& 46\% / 88.5\% / 60.5\%      \\

\multicolumn{1}{l}{\textsc{Easy-IPC-Grid} (30)} & \multicolumn{1}{l}{14.1}         & 48.5         
& 0.52			& \multicolumn{3}{l}{83.9\% / 92.9\% / 88.1\%}         
& 0.30			& 100\% / 96.2\% / 98\%
& 0.31			& 83.9\% / 92.9\% / 88.1\%         \\

\multicolumn{1}{l}{\textsc{Ferry} (30)}         & \multicolumn{1}{l}{13.8}         & 18.1         
& 0.38			& \multicolumn{3}{l}{57.6\% / 67.9\% / 62.3\%}
& 0.13			& 100\% / 78.6\% / 88\%
& 0.15			& 57.6\% / 67.9\% / 62.3\%        \\

\multicolumn{1}{l}{\textsc{Logistics} (30)}     & \multicolumn{1}{l}{20.8}         & 24.0           
& 0.70			& \multicolumn{3}{l}{100\% / 13\% / 23.1\%}
& 0.11			& 100\% / 91.3\% / 95.5\%       
& 0.13			& 100\% / 13\% / 23.1\%       \\

\multicolumn{1}{l}{\textsc{Satellite} (20)}     & \multicolumn{1}{l}{25.7}         & 60.8         
& 3.45			& \multicolumn{3}{l}{75\% / 75\% / 75\%}            
& 0.44			& 45.5\% / 66.7\% / 54.1\%
& 0.50			& 100\% / 26.7\% / 42.1\%      \\

\multicolumn{1}{l}{\textsc{Sokoban} (20)}       & \multicolumn{1}{l}{24}           & 76.5         
& 3.77			& \multicolumn{3}{l}{10.5\% / 42.9\% / 16.9\%}
& 1.33			& 5.4\% / 14.3\% / 7.8\%
& 1.45			& 35.5\% / 78.6\% / 48.9\%     \\

\multicolumn{1}{l}{\textsc{Zeno-Travel} (15)}   & \multicolumn{1}{l}{12.2}         & 38.7         
& 0.71			& \multicolumn{3}{l}{77.8\% / 50\% / 60.9\%}
& 0.14			& 82.4\% / 100\% / 90.3\%       
& 0.18			& 77.8\% / 50\% / 60.9\%       \\ \hline
\end{tabular}
\caption{Experimental results for detecting sub-optimal action steps using the technique Analysing Plan Execution Deviation (2).}
\label{tab:Table2:OptimalityMonitoring-PlanDeviation}
\end{table*}

\begin{table*}[ht!]
\centering
\begin{tabular}{lllllllllll}
\hline
		\multicolumn{11}{c}{\begin{tabular}[c]{@{}c@{}} Plan Optimality Monitoring \\ (Analysing Plan Execution Deviation and Predicting Upcoming Actions via Landmarks) \end{tabular}}	\\ \hline
		&                                   &              & \multicolumn{4}{c}{$h_{\mathit{adjsum}}$}                                     & \multicolumn{2}{c}{$h_{\mathit{adjsum2}}$}            & \multicolumn{2}{c}{$h_{\mathit{adjsum2M}}$}           \\ \hline
\multicolumn{1}{l}{\textbf{Domain}}    & \multicolumn{1}{l}{\textbf{$|O|$}} & \textbf{$|\mathcal{L}|$} & \textbf{Time}    & \multicolumn{3}{l}{\textbf{PPV / TPR / F1}} & \textbf{Time} & \textbf{PPV / TPR / F1} & \textbf{Time} & \textbf{PPV / TPR / F1} \\ \hline

\multicolumn{1}{l}{\textsc{Driver-Log} (20)}    & \multicolumn{1}{l}{20.1}         & 53.6         
{\hskip 12pt} & 0.71             & \multicolumn{3}{l}{100\% / 77.7\% / 87.5\%}       
{\hskip 12pt} & 0.68          & 100\% / 94.4\% / 97.1\%       
{\hskip 12pt} & 1.33          & 100\% / 100\% / 100\%         \\

\multicolumn{1}{l}{\textsc{Depots} (30)}        & \multicolumn{1}{l}{16.7}         & 64.7         
& 1.34             & \multicolumn{3}{l}{71.8\% / 88.4\% / 79.3\%}      
& 1.22          & 81.2\% / 100\% / 89.6\%       
& 2.15          & 75.6\% / 93.3\% / 83.5\%      \\

\multicolumn{1}{l}{\textsc{Easy-IPC-Grid} (30)} & \multicolumn{1}{l}{14.1}         & 48.5         
& 0.81             & \multicolumn{3}{l}{100\% / 96.1\% / 98\%}         
& 0.77          & 100\% / 100\% / 100\%         
& 0.98          & 100\% / 75\% / 85.7\%         \\

\multicolumn{1}{l}{\textsc{Ferry} (30)}         & \multicolumn{1}{l}{13.8}         & 18.1         
& 0.23             & \multicolumn{3}{l}{88\% / 78.5\% / 83.1\%}        
& 0.18          & 88\% / 78.5\% / 83.1\%        
& 0.34          & 80\% / 42.9\% / 55.8\%        \\

\multicolumn{1}{l}{\textsc{Logistics} (30)}     & \multicolumn{1}{l}{20.8}         & 24.0           
& 0.47             & \multicolumn{3}{l}{100\% / 85.7\% / 92.3\%}       
& 0.35          & 100\% / 91.3\% / 95.4\%       
& 0.89          & 100\% / 91.3\% / 95.4\%       \\

\multicolumn{1}{l}{\textsc{Satellite} (20)}     & \multicolumn{1}{l}{25.7}         & 60.8         
& 5.41             & \multicolumn{3}{l}{100\% / 26.6\% / 42.1\%}       
& 4.35          & 87.5\% / 46.6\% / 60.8\%      
& 9.58          & 88.8\% / 53.3\% / 66.6\%      \\

\multicolumn{1}{l}{\textsc{Sokoban} (20)}       & \multicolumn{1}{l}{24.0}           & 76.5         
& 3.45			& \multicolumn{3}{l}{64.7\% / 78.6\% / 71.0\%}
& 2.26          & 80.0\% / 57.1\% / 66.7\%
& 4.13          & 60.0\% / 64.3\% / 62.1\%      \\

\multicolumn{1}{l}{\textsc{Zeno-Travel} (15)}   & \multicolumn{1}{l}{12.2}         & 38.7         & 1.07             & \multicolumn{3}{l}{87.5\% / 50\% / 63.6\%}        & 0.86          & 100\% / 92.8\% / 96.2\%       & 1.52          & 100\% / 85.7\% / 92.3\%       \\ \hline
\end{tabular}
\caption{Experimental results for detecting sub-optimal action steps combining the techniques Analysing Plan Execution Deviation and Predicting Upcoming Actions via Landmarks (1).}
\label{tab:Table1:OptimalityMonitoring}
\end{table*}

\begin{table*}[ht!]
\centering
\begin{tabular}{lllllllllll}
\hline
		\multicolumn{11}{c}{\begin{tabular}[c]{@{}c@{}} Plan Optimality Monitoring \\ (Analysing Plan Execution Deviation and Predicting Upcoming Actions via Landmarks) \end{tabular}}	\\ \hline
		&                                   & \multicolumn{1}{c}{} & \multicolumn{4}{c}{$h_{\mathit{combo}}$}                                          & \multicolumn{2}{c}{$h_{\mathit{ff}}$}             & \multicolumn{2}{c}{$h_{\mathit{sum}}$}                      \\ \hline
\multicolumn{1}{l}{\textbf{Domain}}    & \multicolumn{1}{l}{\textbf{$|O|$}} & \textbf{$|\mathcal{L}|$}          & \textbf{Time}  & \multicolumn{3}{l}{\textbf{PPV / TPR / F1}} & \textbf{Time} & \textbf{PPV / TPR / F1} & \textbf{Time} & \textbf{PPV / TPR / F1} \\ \hline

\multicolumn{1}{l}{\textsc{Driver-Log} (20)}    & \multicolumn{1}{l}{20.1}         & 53.6                  
{\hskip 12pt} & 1.38           & \multicolumn{3}{l}{100\% / 77.7\% / 87.5\%}       
{\hskip 12pt} & 0.74          & 100\% / 94.4\% / 97.1\%       
{\hskip 12pt} & 0.85          & 100\% / 77.7\% / 87.5\%       \\

\multicolumn{1}{l}{\textsc{Depots} (30)}        & \multicolumn{1}{l}{16.7}         & 64.7                  
& 2.46           & \multicolumn{3}{l}{71.4\% / 96.1\% / 81.9\%}      
& 1.43          & 81.2\% / 100\% / 89.6\%       
& 1.39          & 71.8\% / 88.4\% / 79.3\%      \\

\multicolumn{1}{l}{\textsc{Easy-IPC-Grid} (30)} & \multicolumn{1}{l}{14.1}         & 48.5                  
& 1.08          & \multicolumn{3}{l}{100\% / 96.1\% / 98\%}         
& 0.86          & 100\% / 100\% 100\%           
& 0.79          & 100\% / 96.1\% / 98\%         \\

\multicolumn{1}{l}{\textsc{Ferry} (30)}         & \multicolumn{1}{l}{13.8}         & 18.1                  
& 0.36           & \multicolumn{3}{l}{88\% / 78.5\% / 83.1\%}        
& 0.32          & 88\% / 78.5\% / 83.1\%        
& 0.19          & 88\% / 78.5\% / 83.1\%        \\

\multicolumn{1}{l}{\textsc{Logistics} (30)}     & \multicolumn{1}{l}{20.8}         & 24.0                    
& 1.11           & \multicolumn{3}{l}{100\% / 85.7\% / 92.3\%}       
& 0.55          & 100\% / 91.3\% / 95.4\%       
& 0.43          & 100\% / 85.7\% / 92.3\%       \\

\multicolumn{1}{l}{\textsc{Satellite} (20)}     & \multicolumn{1}{l}{25.7}         & 60.8                  
& 9.81           & \multicolumn{3}{l}{100\% / 40\% / 57.1\%}         
& 4.94          & 87.5\% / 46.6\% / 60.8\%      
& 4.53          & 100\% / 26.6\% / 42.1\%       \\

\multicolumn{1}{l}{\textsc{Sokoban} (20)}       & \multicolumn{1}{l}{24.0}           & 76.5                  
& 4.28          & \multicolumn{3}{l}{73.3\% / 78.6\% / 75.9\%}
& 2.22          & 80.0\% / 57.1\% / 66.7\%
& 2.07          & 64.7\% / 78.6\% / 71.0\%            \\

\multicolumn{1}{l}{\textsc{Zeno-Travel} (15)}   & \multicolumn{1}{l}{12.2}         & 38.7                  
& 1.45           & \multicolumn{3}{l}{87.5\% / 50\% / 63.6\%}        
& 0.99          & 100\% / 92.8\% / 96.2\%       
& 0.92          & 87.5\% / 50\% / 63.6\%        \\ \hline
\end{tabular}
\caption{Experimental results for detecting sub-optimal action steps combining the techniques Analysing Plan Execution Deviation and Predicting Upcoming Actions via Landmarks (2).}
\label{tab:Table2:OptimalityMonitoring}
\end{table*}
\end{landscape}
}

\newpage
\subsection{Commitment Abandonment Detection Experiments}
\label{subsection:CommitmentAbandonmentExperiments}

We evaluated our approach to detect commitment abandonment using the same metrics as before: \textit{Precision} (PPV), \textit{Recall} (TPR), and the \textit{F1-score} (F1). 
Here, true positive results represent the number of plans that actually did abandon their expected commitments that our approach has detected correctly. 
False positive results represent the number of plans that actually eventually achieved the commitment consequent that our approach has detected as having abandoned the commitment. 
False negative results represent the number of plans that would not eventually reach the commitment consequent that our approach has not detected as abandonment.

Table~\ref{tab:ExperimentsCommitment} shows the experimental results of our commitment abandonment approach over the selected domains using the heuristics that yield best results to detect sub-optimal action steps. 
Each row details results for a different domain showing averages for the number of observations $|O|$ across problem instances, monitoring time (in seconds), \emph{Precision} -- PPV, \emph{Recall} -- TPR, and \emph{F1-score} -- F1. The average number of observations ($|O|$), ranging between 10.0 and 23.5, indicating that all plans we analyse are non-trivial in complexity.\frm{Re-write this like I did in the previous section.} 

For the \textsc{Driver-Log}, \textsc{Easy-IPC-Grid}, and \textsc{Logistics} domains our approach yields perfect predictions to detect commitment abandonment. Apart from the domains \textsc{Satellite} and \textsc{Sokoban}, that yield poor results (for threshold values 5\% and 10\%), for other domains we have near perfect prediction for detecting commitment abandonment. The results for \textsc{Satellite} and \textsc{Sokoban} are not so that good as for other domains because the commitment abandonment detection is related to our plan optimality monitoring approach, in which has not as good as the results for detecting sub-optimal action steps for the same domains, as we shown in Tables~\ref{tab:Table1:OptimalityMonitoring} and~\ref{tab:Table2:OptimalityMonitoring}. Thus, we can conclude that by using the detection of sub-optimal action steps it is possible to identify accurately commitment abandonment in planning domains.


\begin{table*}[ht!]
\footnotesize
\centering
\begin{tabular}{lllllll}
\hline
{\bf Domain}  & {\bf Heuristic}    & {\bf $|O|$} & {\bf Time} & \multicolumn{1}{c}{\begin{tabular}[c]{@{}c@{}}\textbf{PPV}\\ $\theta$ \textbf{(0\% / 5\% / 10\%)}\end{tabular}}  & \multicolumn{1}{c}{\begin{tabular}[c]{@{}c@{}}\textbf{TPR}\\ $\theta$ \textbf{(0\% / 5\% / 10\%)}\end{tabular}} & \multicolumn{1}{c}{\begin{tabular}[c]{@{}c@{}}\textbf{F1}\\ $\theta$ \textbf{(0\% / 5\% / 10\%)}\end{tabular}} \\ \hline

\multicolumn{1}{l}{\begin{tabular}[c]{@{}c@{}} \textsc{Driver-Log} (30) \end{tabular}} & $h_{\mathit{adjsum2M}}$ & 20.0 & 0.83                & 100\% / 100\% / 100\%              & 100\% / 100\% / 100\%            & 100\% / 100\% / 100\%           \\

\multicolumn{1}{l}{\begin{tabular}[c]{@{}c@{}} \textsc{Depots} (30) \end{tabular}}     & $h_{\mathit{adjsum2}}$ & 18.6  & 1.79              & 100\% / 100\% / 100\%             & 100\% / 100\% / 80.0\%            & 100\% / 100\% / 88.8\%           \\

\multicolumn{1}{l}{\begin{tabular}[c]{@{}c@{}} \textsc{Easy-IPC-Grid} (30) \end{tabular}}   & $h_{\mathit{ff}}$ & 17.3 &    0.95        & 100\% / 100\% / 100\%              & 100\% / 100\% / 100\%            & 100\% / 100\% / 100\%          \\

\multicolumn{1}{l}{\begin{tabular}[c]{@{}c@{}} \textsc{Ferry} (30) \end{tabular}}     & $h_{\mathit{adjsum2}}$ & 13.5  & 0.38                & 100\% / 100\% / 100\%             & 100\% / 80.0\% / 80.0\%            & 100\% / 88.8\% / 88.8\%          \\

\multicolumn{1}{l}{\begin{tabular}[c]{@{}c@{}} \textsc{Logistics} (30) \end{tabular}}     & $h_{\mathit{adjsum2}}$ & 21.0  & 0.56            & 100\% / 100\% / 100\%             & 100\% / 100\% / 100\%             & 100\% / 100\% / 100\%            \\

\multicolumn{1}{l}{\begin{tabular}[c]{@{}c@{}} \textsc{Satellite} (30) \end{tabular}}     & $h_{\mathit{adjsum2M}}$ & 23.5 & 5.4             & 80\% / 100\% / 100\%             & 80\% / 60\% / 60\%            & 80\% / 75\% / 75\%           \\

\multicolumn{1}{l}{\begin{tabular}[c]{@{}c@{}} \textsc{Sokoban} (30) \end{tabular}}     & $h_{\mathit{combo}}$ & 22.8   & 5.2             & 83.3\% / 100\% / 100\%             & 100\% / 60\% / 60\%            & 90.9\% / 75\% / 75\%          \\

\multicolumn{1}{l}{\begin{tabular}[c]{@{}c@{}} \textsc{Zeno-Travel} (30) \end{tabular}}     & $h_{\mathit{adjsum2}}$ & 10.0   & 1.1         & 100\% / 100\% / 100\%             & 80\% / 80\% / 80\%            & 88.8\% / 88.8\% / 88.8\%           \\ \hline
\end{tabular}
\caption{Experimental results for detecting commitment abandonment.}
\label{tab:ExperimentsCommitment}
\end{table*}

\newpage
\section{Related Work}
\label{section:RelatedWork}

To the best of our knowledge, the most recent approach to monitor plan optimality was developed by Fritz and McIlraith~\cite{FritzICAPS2007}. 
This work formalises the problem of monitoring plan optimality by using situation calculus, a logical formalism to specify and reason about dynamical systems.
Fritz and McIlraith seek to determine whether an execution follows an optimal plan, but --- unlike our work --- do not seek to determine which actions are responsible for deviation from an optimal plan. Prior to that, Geib and Goldman~\cite{ProblemsWithElderCare_AAAI2002,AbandonPlansGeib_IJCAI2003} develop a formal model of goal and plan abandonment detection. 
This formal model is based on plan libraries and estimates the probability that a set of observed actions in a sequence contributes to the goal being monitored. 
Unlike our work, which requires no prior knowledge of an agent's plan library, they assume knowledge about possible plan decompositions (i.e., a know-how of all plans to achieve a set of goals) available to each observed agent. 

Siddiqui and Haslum~\cite{Siddiqui_AI_Australia_12} developed an approach that aims to detect deorderable (unordered) blocks of partial plans. According to the authors, a partial ordering implies that whenever two sub-plans are unordered, every interleaving steps from the two forms a valid execution. In this work, Siddiqui and Haslum propose a notion of partial ordering that divides the plan into blocks, such that the action steps in a block may not be interleaved with action steps outside the block, but unordered blocks can be executed in any sequence. The authors argue that this approach can be used, for example, to break plans into sub-plans for distributed executions.

Proposed by Kafal{\i}~et al.~\cite{Kafali_ECAI2014}, \textsc{Gosu} is an algorithm that uses commitments to regulate agents' interactions in an environment for achieving their goals. By using commitments as contractual agreements, this algorithm allows an agent to decide whether it can achieve its goals for a given set of commitments and the current state. \textsc{Gosu} does not use any planning approach to reason about agents' goals, it uses a depth-first search algorithm.

In~\cite{KafaliY_KIS2016}, Kafal{\i} and Yolum propose a monitoring approach called PISAGOR that can determine whether a set of business interactions are progressing as expected in an e-commerce system. These business interactions are represented as commitments with deadlines. The authors also propose a set of operational rules for the observed agent in order to create expectations based on its commitments. Thus, PISAGOR monitors and detects whether the observed interaction is progressing well, and therefore it identifies what the problem is during the interactions.

Our technique to predict upcoming actions via landmarks (Section~\ref{subsection:PredictingUpcomingActions}) is inspired by the landmark-based heuristics developed in~\cite{PereiraMeneguzzi_ECAI2016,PereiraNirMeneguzzi_AAAI2017}. In this work, Pereira, Oren and Meneguzzi developed two goal recognition heuristics that rely on planning landmarks, showing that it is possible to recognise accurately goals using just the concept of landmarks. 

In~\cite{ijcai18_counterplanning}, Pozanco~et al. develop an approach for counter-planning that combines the use of goal recognition, landmarks, and automated planning. The aim of this work is blocking the opponent's goal achievement by recognising opponent's goal, landmark extraction to identify sub-goals that can be used to block goal achievement, and automated planning to generate plans that prevent goal achievement.

Criado~\cite{Criado_Norms_2018} develops a multi-agent system approach to monitor and check norm compliance with limited resources, assuming incomplete information about the performed actions and the state of the world. This approach uses propositional logic to formalise the properties (state) of the world and norms, and action definitions with preconditions and effects to formalise how the agents act in the environment. For experiments and evaluation, this approach uses planning domains and problems (some of which we used in this work) also used in~\cite{RamirezG_IJCAI2009}.

In~\cite{Meneguzzi2018_JAAMAS}, Meneguzzi et al. develop a Hierarchical-Planning formalisation of the interplay between commitments and goals, allowing agents that have a global view of the agent society to decide offline whether certain goals can indeed be accomplished either by individual agents or by complex networks of commitments. 
Such networks allow agents to delegate part of their goals and cooperate towards achieving interdependent individual goals. 
By having a global view of the planning problem and the delegation, Meneguzzi et al. assume that agents that do commit to each other achieve delegated goals in order to decide, offline, whether the goals can be achieved. 
Our work, by contrast, takes an online individualistic view of each agent as it delegates goals to other agents. 
Here, individual agents monitor debtors while they act, and, when such debtors fail to achieve the consequent of a commitment may take appropriate action to ensure they find alternative ways to achieve delegated goals. 

Most recently, Telang et al.~\cite{JAIR_2019_Commitments} formalise and prove a number of properties of the combined goal and commitment protocols implemented by Telang, Meneguzzi and others~\cite{Meneguzzi2018_JAAMAS,Commitments_MeneguzziTS13,CommitmentsHTN_Telang2013}. 
Such formalisation takes a similar offline global view of goals and commitments in a larger society, and focus on similarly global properties of the \emph{possible} plans that can achieve individual agent goals and fulfil their commitments, whereas our work reasons about the actual plans used by the agents as they operate in the environment. 

\section{Conclusions}
\label{section:Conclusions}

In this article, we have developed planning-based approaches for monitoring plan optimality and detecting commitment abandonment.
Our approaches use planning techniques, but do not require executing a full planning algorithm by exploiting landmarks and domain-independent heuristics. 
These provide useful information that can be used in planning-based approaches for recognising goals and plans. 
Finally, our plan optimality approach can also be used to repair sub-optimal plans, for example, by detecting which parts of the plan are sub-optimal, and then improving it.

As we show in experiments and evaluation, our approaches yield very accurate results for detecting sub-optimal plan steps and commitment abandonment, dealing with realistic well-known deterministic planning domains. 
Thus, our approaches can provide timely accurate estimates of when an agent fails to accomplish a delegated commitment, allowing creditors (i.e. the agents that rely on the commitment being achieved) to decide on which action to take to achieve their individual goals. 
Such actions might consist of reminding the debtor or sanctioning unreliable debtors (when applicable) and quickly arrange for an alternative agent to accomplish the desired delegated goal. 
Such techniques thus provide a major contribution to the design of autonomous agent societies that rely on networks of commitments to achieve societal goals~\cite{AAAI-Commit-08,Commitments_MeneguzziTS13,Baldoni:2015fe}, allowing agents to quickly compute which agents are accountable~\cite{Baldoni:2018hm} when networks of commitments fail. 

\subsection{Limitations and Future Work}

The work described in this paper makes several simplifying assumptions, but these assumptions impact on its applicability to some domains. Therefore, one strand of future work involves relaxing these assumptions. One such assumption is of full observability of the environment during the monitoring process. An approach to dealing with partial observability involves using an automated planner to fill in missing observations. Related to this is our assumption of uniform action costs, which we can relax through the use of cost-based planning heuristics~\cite{Keyder_ECAI08_CostBasedHeuristics,Florian_ICAPS2014_CostBasedHeuristics}. Combining these two will introduce a new assumption, namely that the (least cost) plan identified by the planner is the one utilised by the agent. Evaluating whether this new assumption is appropriate is another important strand of work. Finally, to deal with interleaving plans and multiple goals, we will use other landmark extraction algorithms~\cite{ICAPS03_DC_ZhGi,LandmarksRichter_2008,KeyderRH_ECAI10_Landmarks}, and investigate the effectiveness of more modern planning heuristics~\cite{SeippPH15_ICAPS,SieversWH16_ICAPS,PommereningHB17_AAAI}.


\section*{Acknowledgements}

This work is financed by the Coordena\c{c}\~{a}o de Aperfei\c{c}oamento de Pessoal de Nivel Superior (Brazil, Finance Code 001).
Felipe thanks CNPq for partial financial support under its PQ fellowship, grant number 305969/2016-1.

\bibliographystyle{ACM-Reference-Format}
\bibliography{tist-journal-commitments}

\end{document}